\documentclass[10pt,journal,compsoc]{IEEEtran}

\ifCLASSOPTIONcompsoc
  \usepackage[nocompress]{cite}
\else
  \usepackage{cite}
\fi
\usepackage[export]{adjustbox}
\usepackage{array}
\usepackage{enumitem}
\usepackage{graphicx}
\usepackage{mathptmx}
\usepackage{tikz}
\usepackage{times}
\usepackage[hidelinks]{hyperref}
\usepackage{amsmath}
\usepackage{booktabs}
\usepackage{subcaption}
\usepackage{float}
\usepackage{multirow}

\usepackage[none]{hyphenat}
\emergencystretch=\maxdimen
\DeclareMathOperator*{\argmax}{\texttt{arg\,max}}
\DeclareMathOperator*{\argmin}{\texttt{arg\,min}}
\DeclareMathOperator*{\mean}{\texttt{mean}}
\DeclareMathAlphabet{\mathcal}{OMS}{cmsy}{m}{n}

\begin{document}
%
\title{On the Effect of Pre-Processing and Model Complexity for Plastic Analysis Using Short-Wave-Infrared Hyper-Spectral Imaging}
%
%
%
%

\author{Klaas Dijkstra$^1$, Maya Aghaei$^1$, Femke Jaarsma$^2$, Martin Dijkstra$^1$, Rudy Folkersma$^2$, Jan Jager$^2$, Jaap van de Loosdrecht$^1$}

\author{Klaas~Dijkstra,
        Maya~Aghaei,
        Femke~Jaarsma,
        Martin~Dijkstra,
        Rudy~Folkersma,
        Jan~Jager,
        and~Jaap~van~de~Loosdrecht
\IEEEcompsocitemizethanks{\IEEEcompsocthanksitem K. Dijkstra, M. Aghaei, M. Dijkstra, and J. Loosdrecht were with the Professorship Computer Vision \& Data Science, NHL Stenden University of Applied Sciences. 

\IEEEcompsocthanksitem F. Jaarsma, R. Folkersma and J. Jager were with the Professorship Circular plastics, NHL Stenden University of Applied Sciences.\protect\\

Address: Rengerslaan 10, 8917 DD, Leeuwarden, The Netherlands.

E-mail: name.surname@nhlstenden.com}}
\IEEEtitleabstractindextext{%
\begin{abstract}

The importance of plastic waste recycling is undeniable. In this respect, computer vision and deep learning enable solutions through the automated analysis of short-wave-infrared hyper-spectral images of plastics. In this paper, we offer an exhaustive empirical study to show the importance of efficient model selection for resolving the task of hyper-spectral image segmentation of various plastic flakes using deep learning. We assess the complexity level of generic and specialized models and infer their performance capacity: generic models are often unnecessarily complex. We introduce two variants of a specialized hyper-spectral architecture, PlasticNet, that outperforms several well-known segmentation architectures in both performance as well as computational complexity. In addition, we shed lights on the significance of signal pre-processing within the realm of hyper-spectral imaging. To complete our contribution, we introduce the largest, most versatile hyper-spectral dataset of plastic flakes of four primary polymer types.

\end{abstract}

}

\maketitle

\IEEEdisplaynontitleabstractindextext

%
\IEEEpeerreviewmaketitle

\IEEEraisesectionheading{\section{Introduction}\label{sec:introduction}}

%
%
%
%
\IEEEPARstart{P}{lastics} are versatile and cost-effective materials with often single-time usage applications. This leads to the waste ending up in nature and therefore having an increasingly negative impact on our planet \cite{chow2017plastic}. 
The most common plastics in municipal waste are polyethylene (PE), polypropylene (PP), polyethylene terephthalate (PET) and polystyrene (PS), which are all non-degradable in nature. It is estimated that around 3\% of the total plastic production worldwide ends up directly in nature, the majority is landfilled or thermally incinerated \cite{taghavi2021challenges}. Current efforts focus on improving the process of recycling to recover resources. For mechanical recycling, plastics are collected, sorted on polymer type, shredded and washed as a preparation for remelting \cite{voet2021plastics}. However, since objects in waste usually contain multiple polymer types, the analysis and subsequently sorting of these polymers is important.

This paper focuses on analyzing the polymer types of real waste products which have been shredded and placed on a moving conveyor belt to simulate the practical use-case. We use Hyper-Spectral Imaging (HSI) for differentiating polymer flakes. The difference between polymer types can be observed using Short-Wave Infrared (SWIR) cameras due to their absorption characteristic of the electromagnetic spectrum. Detection with SWIR cameras requires little or no sample preparation and can be done near real-time, which is an advantage over typical analytical methods. Specific molecular bonds that are present in polymers react differently to the specific electromagnetic wavelengths. Absorption peaks in the spectrum correspond to vibrations of the specific molecular bonds, this gives each type of polymer a specific spectral profile. The camera used in our experiments is only able to perceive overtones of the absorption peaks, which are less pronounced, and therefore require more effort to be used as a distinguishing signal \cite{eldin2011near}. This is the preferred method in our research because cameras that are able measure the fundamental frequencies are an order more expensive.


A traditional method of spectral analysis is using the linear methods to correlate the known spectra of a polymer with an unknown spectra, using for example Spectral Angle Mapping (SAM) \cite{panda2015hyperspectral}. However, a limitation of (linearly) classifying individual spectral pixels of images is the loss of context information. The category of a pixel can be estimated more accurately by using several neighboring pixels of it \cite{king2006pixel}. Convolutional layers of deep neural networks achieve this by using a convolution filter that is trained to extract the right features from the pixel neighborhood to improve the per-pixel classification performance, or in deep learning jargon, the segmentation result. By using multiple convolutional layers in combination with max-pooling and Rectified Linear Units (ReLU), deep learning approaches have shown impressive results for several segmentation tasks, including the segmentation of Hyper-Spectral (HS) data \cite{sun2021supervised}. Another advantage of deep learning models is that they are able to capture non-linear data patterns which gives them a clear advantage over linear models like SAM.

The main limitations of such data-driven algorithms is that they require large annotated datasets for learning. For these models to generalise well after training, the dataset needs a data distribution of the training data which must be similar to the distribution of the real-world data. In practice the real-world data is simulated by using a held-out dataset for testing. A secondary limitation is the computational complexity of these models. Existing models are generic and have a large number of parameters, so to be able to learn complex patterns in large datasets regardless of the problem. However, such aspects of generic models, often considered to be positive, might in practice be sub-optimal. For less-demanding applications, the complexity of the models could be reduced to prevent overfitting and to, simultaneously, reduce computational complexity. This is explored by the question \emph{What is the effect of computational complexity on the models' performance?}. Another method is to reduce the complexity of the input signal by removing unnecessary information or mapping it to a different space. Various pre-processing methods exist to achieve this effect which are addressed by the following question: \emph{"How should the HS data be pre-processed to to improve the model' performance?"} In this paper and, in the attempt to provide an answer to these questions, we provide an ablation study of various pre-processing algorithms and explain their relation to the classic linear methods. Secondly, we perform an ablation study of several models with varying complexity and provide a narrative to explain the logic behind the design choices. Additionally, we devise a customly-designed neural network called SAMNet which is a trainable version of the SAM algorithm. Moreover, we introduce a simple architecture called PlasticNet and compare it within various studies with the popular U-Net \cite{ronneberger2015u} model. The models and pre-processing methods are tested on a large HS dataset containing PE, PP, PET and PS flakes with entirely held-out evaluation test sets. 

\subsection{Related work}

An HS image contains reflections from the objects in many equally spaced bands of the electromagnetic spectrum, where different objects exhibit their unique transmission and absorption characteristic, defined as their HS signature. HS signature of an object, given beyond visible spectrum, offers a unique potential for deeper analysis of the object. In this regard, the classical machine learning algorithms tend to score low in their analysis \cite{ozdemir2020deep}, the main problem remaining to be the curse of dimensionality, related to the number of channels in an HS image. A common approach is to employ a linear model for HS image analysis such as SAM \cite{calin2018comparison}  or Partial Least Squares (PLS) \cite{uzair2013hyperspectral} to determine the similarity of an unknown spectral signature with a known one. In such approaches, the high dimensionality of the HS data is normally pre-processed to a lower dimension using dimensionality reduction techniques such as PCA \cite{li2017cell}, ICA \cite{wang2006independent}, (multi) band selection \cite{dijkstra2017hyper}, etc. The authors of \cite{dijkstra2017hyper} tested several combination of linear pre-processing and non-linear patch-wise classification techniques on HS images and demonstrated the suitability of their proposed technique for their specific case study. The authors also observed that increasing the patch size, leads to improvements in performance, indicating the benefits of using the spatial information around pixels for deep learning based classifiers compared to linear models. In a recent work \cite{li2021research}, the authors introduce HS image segmentation, correction and spectral-spatial denoising as HS image pre-processing to improve the HS image classification accuracy.

While the performance of classical machine learning algorithms is largely dependent on the quality of (hand-crafted) reduced features \cite{tarabalka2010segmentation, li2017cell}, deep learning based architectures have shown ever-increasing success for HS image analysis thanks to the automatic and hierarchical feature learning process directly from the data which leads to a model with various semantic layers suitable for representing the computer vision task at hand (e.g., classification, segmentation, detection, etc.) \cite{signoroni2019deep}. Initially, most studies investigated single pixel wise role of spectral signatures of images \cite{xing2016stacked}. In later studies, the joint analysis of spatial-spectral properties of HS image revealed its importance for improving the performance \cite{imani2020overview}. Convolutional Neural Networks (CNNs) as the most popular approach in computer vision, facilitate spatial-spectral joint exploitation thanks to their convolutional filters through the inclusion of neighboring pixels for analysis of a region \cite{zhong2017spectral, yang2017learning}. In the same line, Sun et al. \cite{sun2019spectral} proposed spatial-spectral attention network (SSAN) to capture discriminative spatial-spectral features from attention areas of HS cubes. In addition to CNNs, some other efforts have been dedicated to extract spectral features using Deep Belief Networks (DBNs) \cite{zhong2017learning} and
Recurrent Neural Networks (RNNs) \cite{mou2017deep}. In \cite{nalepa2019validating}, the authors proposed a novel strategy towards unbaised validation of segmentation algorithms for HS images by preventing leakage of data from training set to validation sets. Their proposal although insightful and essential for currently existing HSI datasets, does not necessarily apply to our case thanks to the ample size of our proposed dataset in this study. In fact, one of the contributions of this work revolves around the introduction and exploration of an HSI dataset consisting of independently recorded images for the purpose of training, validation and testing. We exploit the potentials of this dataset to further investigate the algorithmic complexity for pixel-level classification in HS images, introducing new and adapting existing segmentation algorithms for this purpose.

The exclusive potential of HSI hence justify its vast range of applications. Among such problems, is identification of plastics. The authors in \cite{serranti2018characterization} introduce a classification method employing Partial Least Square Discriminant Analysis (PL-SDA) and PCA for recognition of marine microplastic litter of three polymers: PE, PP and PS. In another work \cite{bonifazi2018hierarchical}, the authors recognize different polymer flakes focusing on high-density versus low-density PE polymers from mixed plastic waste through a hierarchical PL-SDA classification strategy. Zheng et al. \cite{zheng2018discrimination} propose a Fisher discrimination model for identification of six groups of plastics: PS, PP, PE, PET, Acrylonitrile Butadiene Styrene (ABS) and PolyVinyl Chloride (PVC), resulting in 100\% prediction of unseen samples. This performance although promising, is achieved on a small unseen set, therefore does not confirm the generalizability. In this work, our primary focus is to shed light on the importance of spectral and spatial pre-processeing of HS images for effective processing of four plastics: PE, PP, PS and PET. We demonstrated this importance within the context of flake segmentation. 

\section{Polymer Hyper-Spectral Dataset}

Due to the lack of an ample and publicly available HS datasets for analysis of polymer flakes, we opted for generating our own dataset, namely Polymer Hyper-Spectral (PolymerHS) dataset\footnote{Dataset will be publicly available upon the publication of the manuscript.}. PolymerHS contains multiple samples per four of the common plastics: PE, PP, PS and PET (see Table \ref{table:class-population}). The samples originate from urban waste, collected, washed and shredded into flakes. Each sample is selected to maximize the diversity of the dataset in terms of original use of the sample, color, thickness, brilliance, and opacity. 

\begin{figure}
    \centering
    \includegraphics[width=7cm]{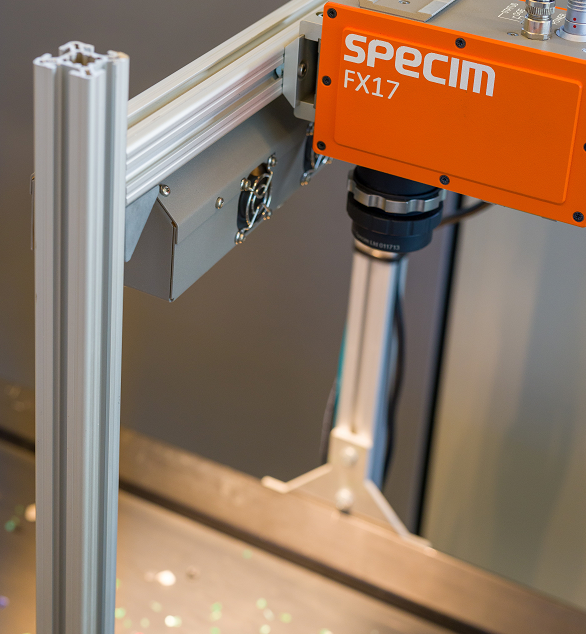}
    \caption{PolymerHS dataset acquisition setup}
    \label{fig:fx17setup}
\end{figure}

The PolymerHS consists of three subsets: Baseline, Test and Test-Dark (the latter acquired at one-third of the camera exposure value used for the Baseline and Test subset acquisition). Baseline subset contains three independently acquired HS images per each pre-chosen sample for this subset. The three images essentially differ in random placement of the same flakes over the conveyor belt. In the Baseline subset, each of the three images per sample is assigned to either train, validation or test, leading to a same size for all parts. In fact, the Baseline subset is designed to provide a reference point for assessing the impact of each experiment, excluding the probable deviations in real world scenarios. On the other side, the Test and Test-Dark subsets, are designed to verify the experiments performances in a more similar setting to the real world scenarios. Five samples per polymer are chosen for creation of the Test and the Test-Dark subsets, with the same sample selection criteria considered for the Baseline subset. To maximize the diversity further, all the samples used for the Test and Test-Dark subsets are selected different from the existing samples for the Baseline subset. The Test-Dark subset particularly adapts a clear distribution outside of the distribution of the original training and validation (part of the Baseline subset), hence will be used to investigate the ultimate capacity of various pre-processing methods and models to extrapolate from the training data.  

\begin{table*}[htbp]
\footnotesize
\centering
\begin{tabular}{@{}llrr|rrr|rrr|rrrr@{}}
\toprule
    & \multicolumn{3}{c}{PE}  & \multicolumn{3}{c}{PP}  & \multicolumn{3}{c}{PS}  & \multicolumn{3}{c}{PET} \\ \cmidrule(l){2-13} 
                           & Images & Flakes & Pixels  & Images & Flakes & Pixels  & Images & Flakes & Pixels  & Images  & Flakes & Pixels \\ \cmidrule(r){1-13}
Baseline (Train+Val)                   & 22   & 802   & 710,758 & 40   & 1,813   & 1,162,635 & 22   & 2,718   & 1,573,171 & 16    & 586    & 495,910 \\
Baseline (Test)                  & 11   & 390   & 333,384 & 20   & 912   & 589,996 & 11   & 1298   & 786,706 & 8    & 310    & 249,076 \\
Test                       & 5    & 127    & 112,986  & 5    & 123    & 86,113   & 5    & 188    & 102,868  & 5     & 126    & 142,575 \\
Test-Dark                  & 5    & 108    & 92,676   & 5    & 118    & 86,513   & 5    & 142    & 76,335   & 5     & 100    & 129,951 \\ \bottomrule
\end{tabular}
\caption{Polymer-instance population in PolymerHS.}
\label{table:class-population}
\end{table*}

This dataset is acquired by capturing the HS signature of plastic flakes randomly scattered over a conveyor belt passing under the Specim FX17 camera\footnote{https://www.specim.fi/products/specim-fx17/}. The illumination is a array of three cooled standard Halogen lights (see Figure~\ref{fig:fx17setup}). The type of camera used in our experiment covers 224 spectral bands between 900 nm and 1700 nm with a fwhm of approximately 3.5 nm. This camera perceives the overtones of the absorption peaks of common polymers\footnote{The cameras capable of seeing the actual peaks, around 5 $\mu$m, are largely less affordable.}. A white reference was used, which has a consistently high diffuse reflectance over the entire spectral range. The dark reference was captured by completely closing the aperture of the camera. These HS images are denoted by $\mathbf{B}^{raw}$, $\mathbf{D}^{raw}$ which are the bright reference and the dark reference image, respectively. Because a line-scan camera is used, both reference images can be averaged over each column in the HS cube using the following equation \cite{zhao2018evaluation}:

\begin{align}
    \mathbf{B}_{x,c} & = \mean_{y \in [h]}(\mathbf{B}^{raw}_{y,x,c}) \\
    \mathbf{D}_{x,c} & = \mean_{y \in [h]}(\mathbf{D}^{raw}_{y,x,c})
\end{align}
, where  $h$ is the height of the HS cube, $y$, $x$, and $c$ are the y-coordinate, x-coordinate and channel , respectively. $\mathbf{B}^m_{x,c}$ is the mean value in a column of the cube $\mathbf{B}$ on location $x$ for channel $c$ and $\mean$ is the mean operator. A corrected image $\mathbf{I}$ is calculated from a raw cube $\mathbf{I}^{raw}$ by:

\begin{align}
    \mathbf{I}_{y,x,c} = \frac{\mathbf{I}^{raw}_{y,x,c} - \mathbf{D}_{x,c}}{\mathbf{B}_{x,c} - \mathbf{D}_{x,c}}
\end{align}

Every instance in the dataset is annotated using Supervisely \footnote{https://supervise.ly/} tool, by defining the boundaries of every flake with a polygon, and assigning it to the corresponding polymer class.

\section{Models}
Information about the type of polymer for plastic flakes is contained in the light-absorbance of specific spectral frequencies. This is captured by the HS signature (See Figure~\ref{fig:spectra}, leftmost). However, the HS signatures might be of varying quality throughout the image due to the noise often caused by the reflections and shadows on the flakes. For example, over-illuminated specular reflections cause the spectra to appear almost full-white, which leads to the loss of most of the spectral information in those pixels. In such case, a model that uses a larger convolutional footprint (i.e. a wider model) is expected to perform better as these models can learn to reduce noise or interpolate lost information using the surrounding pixels. 

\begin{table}[H]
\centering
\begin{tabular}{lrr}
\toprule
{} &  Operations &  Parameters \\
Model        &             &             \\
\midrule
SAMNet       &        1,120 &         1,125 \\
SAMNet3x3    &       10,080 &        10,085 \\
MLPNet       &      108,800 &       109,285 \\
PlasticNet   &      459,564 &       459,765 \\
PlasticNetXL &      798,252 &       798,453 \\
U-Net        &      960,832 &    31,159,301 \\
\bottomrule
\end{tabular}
\caption{Model complexity is defined in two ways: the number of operations per pixel (computational complexity) and the number of parameters (memory size and model capacity)}
\label{tab:complex}
\end{table}

Independent of the presence of noise, shallow models might not be able to capture the spectral signature correctly to classify the polymer types. In fact, adding extra convolutional layers with non-linear activation functions proved to improve the performance \cite{chen2014deep}. Considerably deeper models, using more advanced units such as skip-connections, often demonstrated further improvements for more generic problems, such as medical image segmentation \cite{ronneberger2015u}. In this paper, we aim to see whether this improvement holds up in a more specific scenarios, such as HS image segmentation.

As it is known that most information about the polymer type is contained in the spectral information \cite{voet2021plastics} we investigate where the performance gain is largest and what model complexity is needed for optimal performance. In common sense, the complexity of a model is the byproduct of its width and depth. More specifically, we define model complexity in two ways as seen in Table~\ref{tab:complex}: the number of convolution operations per output pixel and the number of parameters in the model. Hence, as the baseline for a less-complex model, we define SAMNet and, the related but wider SAMNet3$\times$3 which has a footprint of 3$\times$3 pixels. Furthermore, we define MLPNet which has a small footprint (1$\times$1), but has a depth of three layers. Therefore, SAMNet3$\times$3 can be used to observe the added performance for wider models and MLPNet to observe the performance gain for deeper models. The combination of wide and deep is investigated with PlasticNet and PlasticNetXL (with a computational complexity of 0.469 million and 0.798 million operations, respectively). Finally, a standard U-Net is used to investigate if extra performance can be gained from a more complex models with vastly larger number of parameters (over 31 million). Note that, although U-Net has more parameters, the number of operation per pixel is more in line with the PlasticNets (0.96 million operations), this stems from the fact that U-Net uses max-pooling layers which causes the data to be strongly compressed towards the deeper layers of the model, and thus requiring less operations per pixel. In the following subsections the newly introduced models are explained in detail.

\subsection{SAMNet}
Spectral Angle Mapping (SAM) is a linear algorithm that can be used to classify vectors that contain spectral intensities. At the core of the algorithm is the angle between two vectors calculated as:

\begin{align}
    \phi(\mathbf{x}, \mathbf{y}) &= \texttt{cos}^{-1} \left(\frac{\mathbf{x} \cdot \mathbf{y}}{|\mathbf{x}| \times |\mathbf{y}|} \right) \label{eq:sam}
\end{align}
, where $\mathbf{x}$ and $\mathbf{y}$ are two vectors with norms $|\mathbf{x}|$ and $|\mathbf{y}|$, $\cdot$ is the dot product operator and $\phi(\mathbf{x}, \mathbf{y})$ is the minimum angle between the vectors.

Vectors that contain the reference spectra for each type of material that need to be classified are stored as $\mathbf{r} \in \mathcal{R}$. Usually these spectra are initialized as the average spectrum of spectral pixels in a training set, so for example $\mathbf{r}^{\textsc{pet}}$ contains the average spectrum for the PET material. An inference on input HS cube $\mathbf{I}$ is then performed by:

\begin{align}
    \mathbf{C}_{y,x} &= \argmin_{l \in \mathcal{L}}(\phi(\mathbf{I}_{y,x}, \mathbf{r}^l))
    \label{eq:samc}\\
    \mathcal{L} &= \{\textsc{pe}, \textsc{pet}, \textsc{ps}, \textsc{pp}\} 
\end{align}
, where $\mathcal{L}$ is the set of labels, $\mathbf{I}_{y,x}$ is a spectral pixel at location y,x in the input, and $\mathbf{C}$ is the inferred class on that location in the HS cube. The label with the smallest angle to the reference sample is returned by Equation~\eqref{eq:samc}. We propose to translate the SAM algorithm into the CNN paradigm and refer to it as SAMNet. 

In SAMNet, the dot product of Equation~\eqref{eq:sam} is rewritten as a convolutional operator with filter size $h \times w \times c$, where $h$ and $w$ are the height and width, and $c$ is the number of channels of the HS cube. For each class, such a  convolutional filter is initialized with the reference spectra, $\mathbf{r}$, as shown in Equation~\ref{eq:trainsamnet}.

\begin{align}
    \mathbf{F}^{l}_{y,x,i} &= \mathbf{r}^{l}_i \label{eq:trainsamnet} \\
    \mathcal{F} &= \{\mathbf{F}^{l} ~|~ l \in \mathcal{L} \}
\end{align}
, where $\mathbf{r}^{l}_i$ is the $i$th element of the reference spectrum of the class with label $l$. When referring to SAMNet, the size of each convolutional filter is $1 \times 1 \times c$, and consequently, the $y$ and $x$ are always 0. In SAMNet3$\times$3, the filter size is $3 \times 3 \times c$ resulting in a wider model. $\mathcal{F}$ is the filter bank that contains the convolutional filters for each of the classes. In deep learning terminology, this filter bank is simply called a convolutional layer. Similar to Equation~{\eqref{eq:samc}} the inference is performed by:

\begin{align}
    \mathbf{L} &= \mathbf{I'} \otimes \mathcal{F} \\
    \mathbf{C}_{y,x} &= \argmin_{c \in [n]}(\mathbf{L}_{y,x,c})
\end{align}
, where $\mathbf{I}'$ is the normalized HS cube, $\otimes$ is the convolution operator, $\mathbf{L}$ contains the class-logits output, $n$ is the number of classes and $\mathbf{C}_{y,x}$ is the inference result at location $y,x$ where each element indicates the predicted class (the index of the most similar reference spectra $\mathbf{r}$).

To take into account the divisor of Equation~\eqref{eq:sam}, i.e. the division by the multiplied norms, the norm of the convolutional filters and the norm of the spectral dimension of each pixel in the HS cube needs to be normalized to 1 so that the division in Equation~\eqref{eq:sam} can be ignored. 

Firstly, the cube is normalized by dividing each element of the raw input cube $\mathbf{I}$ by the norm of that spectral pixel:

\begin{align}
    \mathbf{I'}_{y,x,c} = \frac{\mathbf{I}_{y,x,c}}{|\mathbf{I}_{y,x}|} \label{eq:sn}
\end{align}
, where $|\mathbf{I}_{y,x}|$ is the norm of the spectral pixel at location $y,x$, and $\mathbf{I}_{y,x,c}$ is an element of the HS cube, and $\mathbf{I}'$ is the spectrally-normalized value. Secondly, the convolution weights can be normalized by dividing each element of each of the convolution filters in the filter bank $\mathcal{F}$ by their norm:

\begin{align}
    \mathbf{F}_{y,x,i} = \frac{\mathbf{F}_{y,x,i}}{|\mathbf{F}_{y,x}|} &&\forall \mathbf{F} \in \mathcal{F}
\end{align}
, where $|\mathbf{F}_{y,x}|$ is the norm of the weights in the convolution filter at location $y$, $x$, and $\mathbf{F}_{y,x,i}$ is an element of the convolution filter. Note that, for SAMNet, both $y$ and $x$ are always 0, and they are both in range [0..2] for SAMNet3$\times$3. The $\texttt{cos}^{-1}$ operator of Equation~\eqref{eq:sam} does not need to be included because is has no effect on the $\argmin$ operator in Equation~\eqref{eq:samc}.

This CNN definition of the SAM algorithm provides more flexibility in terms of changing the footprint and training convolutional parameters\footnote{We experimented with training the SAMNet using deep learning, but found no improvement compared to the SAMNet whose weights were initialized with reference spectra. However, we still think it might be an interesting feature to have for other applications.}.

\subsection{MLPNet}
MLPNet is the CNN equivalent of the Multi-Layer Peceptron (MLP) which is suitable for HS data. It behaves as if running an MLP on each spectral pixel. By formulating this as a CNN, the algorithm is efficient and can be easily implemented using deep learning primitives like convolutional layers. It contains three convolutional layers: the input-, hidden- and output-layer ($\mathcal{I}$, $\mathcal{H}$ and $\mathcal{O}$). In this model, we choose the hidden layer to contain 256 convolutional filters, which is the power of two that is closest to the number of input channels (224). The rationale behind this is to keep the information after the first layer and allow patterns between spectral bands to be captured by the model. Data reduction is achieved in the output-layer.

\begin{align}
    \mathcal{I} &= \{\mathbf{F}_{1 \times 1 \times 224}^{0}, \mathbf{F}_{1 \times 1 \times 224}^{1}, ..., \mathbf{F}_{1 \times 1 \times 224}^{224}\} \\
    \mathcal{O} &= \{\mathbf{F}_{1 \times 1 \times 224}^{0}, \mathbf{F}_{1 \times 1 \times 224}^{1}, ..., \mathbf{F}_{1 \times 1 \times 224}^{256}\} \\
    \mathcal{H} &= \{\mathbf{F}_{1 \times 1 \times 256}^{0}, \mathbf{F}_{1 \times 1 \times 256}^{1}, ..., \mathbf{F}_{1 \times 1 \times 256}^{n}\}
\end{align}
, where $\mathcal{I}$ is a bank with 224 filters, $\mathcal{H}$ is a bank with 256 filters, $\mathcal{O}$ is a bank with $n$ filters, which is equal to the number of output classes. This is equivalent to an MLP with 224 input neurons, 256 hidden neurons and $n$ output neurons, that is applied to each HS pixel. The forward pass of the model is defined by multiple convolutions with ReLU activation functions. The forward step of the MLPNet model is defined as:

\begin{align}
    \mathbf{H}^{mlp1} &= \omega(\mathbf{I} \otimes \mathcal{I}) \\
    \mathbf{H}^{mlp2} &= \omega(\mathbf{H}^{mlp1} \otimes \mathcal{H}) \\
    \mathbf{O}^{mlp} &= \psi(\mathbf{H}^{mlp2} \otimes \mathcal{O})
\end{align}
, where $\mathbf{I}$ is the input HS cube, $\mathbf{H}$ are the intermediate outputs, $\mathbf{O}$ is a tensor containing classification probabilities for each class for each HS pixel in the input, $\omega$ is the TanH activation function, $\psi$ is the Softmax function, and $\otimes$ is the convolution operator. The final classification output is achieved by an $\argmax$ over the spectral dimension.

\subsection{PlasticNet}
PlasticNet is an extension of the MLPNet with a larger footprint and more parameters. We define two architectures which both have three layers, but vary in convolutional footprint. This architecture is designed specifically for HS data. It does not use max-pooling, which is common in many architectures, so it can retain as much spectral information as possible through several layers. Our hypothesis is that these tailor-made, less complex, architectures can outperform more general architectures. The default PlasticNet architecture is defined by halving the number of channels, starting from 224 and also linearly increasing the height and width of the filters, except for the last layer, which only maps to the number of classes using a $ 1\times 1$ convolutional filter.

\begin{align}
    \mathcal{P}^{1} &= \{\mathbf{F}_{3 \times 3 \times 224}^{0}, \mathbf{F}_{3 \times 3 \times 224}^{1}, ..., \mathbf{F}_{3 \times 3 \times 224}^{112}\} \\
    \mathcal{P}^{2} &= \{\mathbf{F}_{5 \times 5 \times 112}^{0}, \mathbf{F}_{5 \times 5 \times 112}^{1}, ..., \mathbf{F}_{5 \times 5 \times 112}^{56}\} \\
    \mathcal{P}^{3} &= \{\mathbf{F}_{7 \times 7 \times 56}^{0}, \mathbf{F}_{7 \times 7 \times 56}^{1}, ..., \mathbf{F}_{7 \times 7 \times 56}^{28}\} \\
    \mathcal{P}^{4} &= \{\mathbf{F}_{1 \times 1 \times 28}^{0}, \mathbf{F}_{1 \times 1 \times 28}^{1}, ..., \mathbf{F}_{1 \times 1 \times 28}^{n}\}
\end{align}
$\mathcal{P}$ are the filter banks of convolutional filters with 112, 56 and 28 filters of size $3 \times 3 \times 224$, $5 \times 5 \times 112$, $7 \times 7 \times 56$ and $1 \times 1 \times 28$ , $n$ is the number of output classes, The forward pass of the model is defined by multiple convolutions with ReLU activation functions.

The forward step of PlasticNet is defined by:

\begin{align}
    \mathbf{H}^{pn1} &= \phi(\mathbf{I} \otimes \mathcal{P}^{1}) \\
    \mathbf{H}^{pn2} &= \phi(\mathbf{H}^{pn1} \otimes \mathcal{P}^{2}) \\
    \mathbf{H}^{pn3} &= \phi(\mathbf{H}^{pn2} \otimes \mathcal{P}^{3}) \\
    \mathbf{O}^{pn} &= \psi(\mathbf{H}^{pn3} \otimes \mathcal{P}^{4})
\end{align}
, where $\mathbf{I}$ is the input HS cube, $\mathbf{H}$ are the intermediate outputs, $\mathbf{O}$ is a tensor containing classification probabilities for each class for each HS pixel in the input, $\phi$ is the ReLU activation function, $\psi$ is the Softmax function, and $\otimes$ is the convolution operator. 

The PlasticNetXL model is similar to the default PlasticNet model but uses convolutional filters of size $3 \times 3 \times 224$, $7 \times 7 \times 112$, $13 \times 13 \times 56$ and $1 \times 1 \times 28$. Here the filters roughly double in size, which should give insight into the effect of using a wider models with more parameters.

\section{Pre-processing}
In a typical deep learning experimental set-up, a common trick is to apply data-augmentation to enrich the training dataset, which is known to also improve the generalization capabilities of the models. In this work, we propose to reach this generalization goal with the opposite approach with various pre-processing methods: instead of enriching the signals, degrading them so to primarily contain relative intensities. By systematically filtering the absolute spectral signal intensities, our hypothesis is that models will learn to focus on relative differences of signals instead, which in turn leads to better generalization.


\begin{figure*}[htbp]
	\includegraphics[width=18cm]{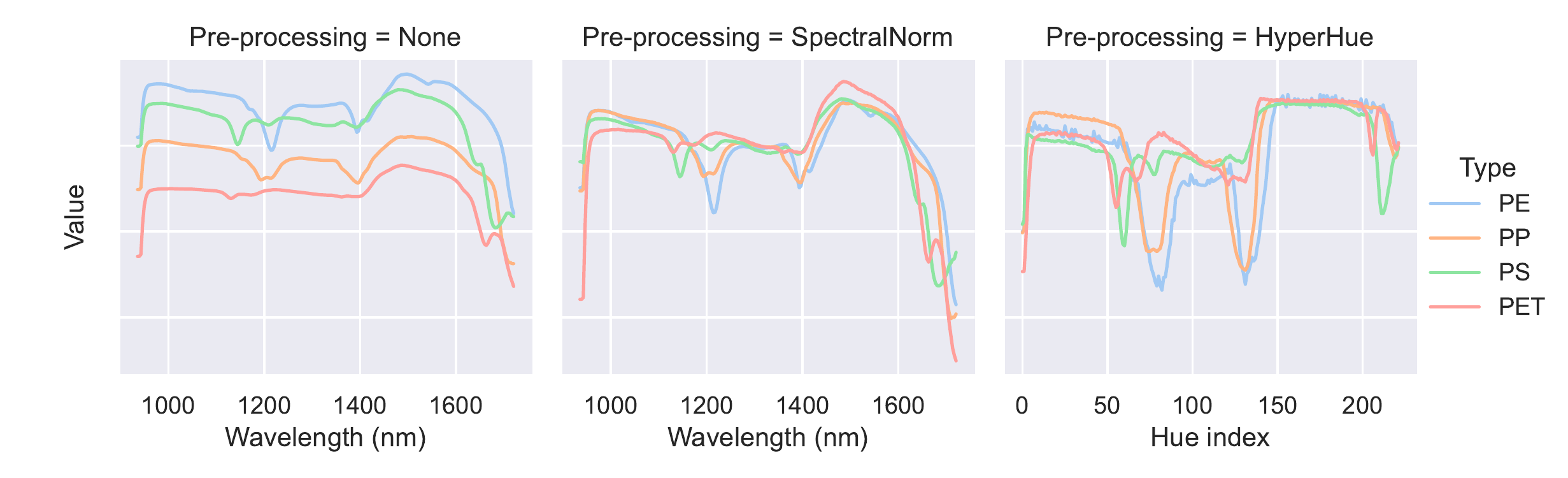}
	\caption{Average value for each polymer type for None, SpectralNorm and HyperHue, calculated for Baseline-Train subset.}
	\label{fig:spectra}
\end{figure*}

\subsection{Derivatives}
Several types of derivatives can be used for pre-processing the HS data. We propose to experiment with derivatives as they are commonly used for finding spectral peaks in data analysis, either manually or with traditional models \cite{bonifazi2018hierarchical}. Derivatives emphasize the relative spectral response of the individual bands. The first derivative $\mathbf{v}'_i = \mathbf{v}_i - \mathbf{v}_{i-1}$ and the second derivative $\mathbf{v}''_i = \mathbf{v}'_i - \mathbf{v}'_{i-1}$, where $\mathbf{v}$ is a spectral vector, have the disadvantage that they preserve the absolute difference in the signal. In such case, the model still runs the risk of over-fitting on the absolute intensity of the signal. On the other side, the logarithmic derivative defined as $\mathtt{ ld}(\mathbf{v}_i) = (\mathbf{v}_i / \mathbf{v}_{i-1}) - 1$, will only preserve the relative fractional difference. As can be seen from the Equation, absolute differences do not cause the output to be different. So, $\mathtt{ld}(\mathbf{v} \times c) == \mathtt{ld}(\mathbf{v})$ for any constant value of $c$,  i.e., if a pixel receives twice the amount of light, and the response of the camera is linear, the logarithmic derivative stays the same. This theoretically shows the method by which the spectral signal is filtered to be less sensitive to varying illumination conditions. The same property is also true for SpectralNorm and HyperHue pre-processing methods introduced in the next subsection.

\subsection{Normalization}
In addition to derivatives which may lead to numerically less stable representation of HS data and in turn complicate model training, we also propose to use normalization. The first proposal is the spectral norm inspired by the SAM algorithm. Equal to Equation~\eqref{eq:sn}, we define the spectral normalization as $\texttt{sn}(\mathbf{v}_i) = \mathbf{v}_i / |\mathbf{v}|$. This normalizes each HS pixel vector $\mathbf{v}$ to a length of 1, making all intensities of all HS pixels equal.

A more sophisticated method of normalization is using the HyperHue \cite{liu2017transformation}. This method can be considered as a generalisation of the HSV color space for HS data. The HS space is projected on the hyper-plane that is perpendicular to the main diagonal. This diagonal represents the axis in which the contribution of all spectral intensities is equal (running from black to white). As a result, the hyper-plane (the space of Hues) is independent of intensities. Additionally, the Saturation channel represents the purity of the Hue and the Value channel represents the intensity of the HS pixel. In our implementation we use the polar representation for the hues instead of the Cartesian representation used in the original paper. This better matches the original definition of the hue in the HSV color space. In the experiments we will evaluate the performance of models with using only Hues and with using the full HyperHSV space.

In Figure~\ref{fig:spectra}, the effect of each of the normalizing pre-processing methods is shown. The left graph shows the average spectra for each of the polymer types without pre-processing. In the center graph the effect of the spectral normalization is shown, where it can be seen that the spectra of the different polymer types overlap and relative differences are mainly visible. In the right graph, the average spectra for the HyperHue pre-processing method are is shown. An interesting observation is that the hyper hues also attempts to normalize between different wavelengths which can be observed from the fact that the hills and crests are of a similar size, consequently the HyperHue method causes more noise.

\section{Experiments \& Results}
In this section, the descriptions and results of the experiments performed using all the aforementioned models and pre-processing methods using PolymerHS dataset is discussed. The training, validation and testing took nearly 60 days of continuous GPU computation. In Table~\ref{tab:results}, the average IoU, precision and recall (taken over all classes) are shown. In this table, several combinations of pre-processing methods and models are compared. For all the experiments, the same Baseline-Train and Baseline-Val subsets were used for training and validation. The models were trained using the Adam optimizer and the hyper-parameters were tuned for each model separately. Each HS cube was randomly tiled during training, with a tile size of $256 \times 256$ to create a rich training dataset. During testing, the HS cube was tiled using a fixed grid of overlapping tiles and later recombined by only using the non-overlapping central part of each tile. This tiling with overlap helps to prevent border effects caused by the footprint of the segmentation models; a similar method was used by \cite{dijkstra2021centroidnetv2}.

For testing, the Baseline-Test is used to shows the baseline performance for each of the model and pre-processing combinations. Furthermore, the Test dataset shows how the performance differs in a more realistic situation and, finally, the Test-Dark dataset is used to shows how the various model and pre-processing combinations respond to images with lower exposure and to determine if a decrease in performance of a model can be mitigated by using pre-processing.

\begin{figure}[htbp]
\centering
    \centering
    \subfloat{{\includegraphics[width=0.95\linewidth]{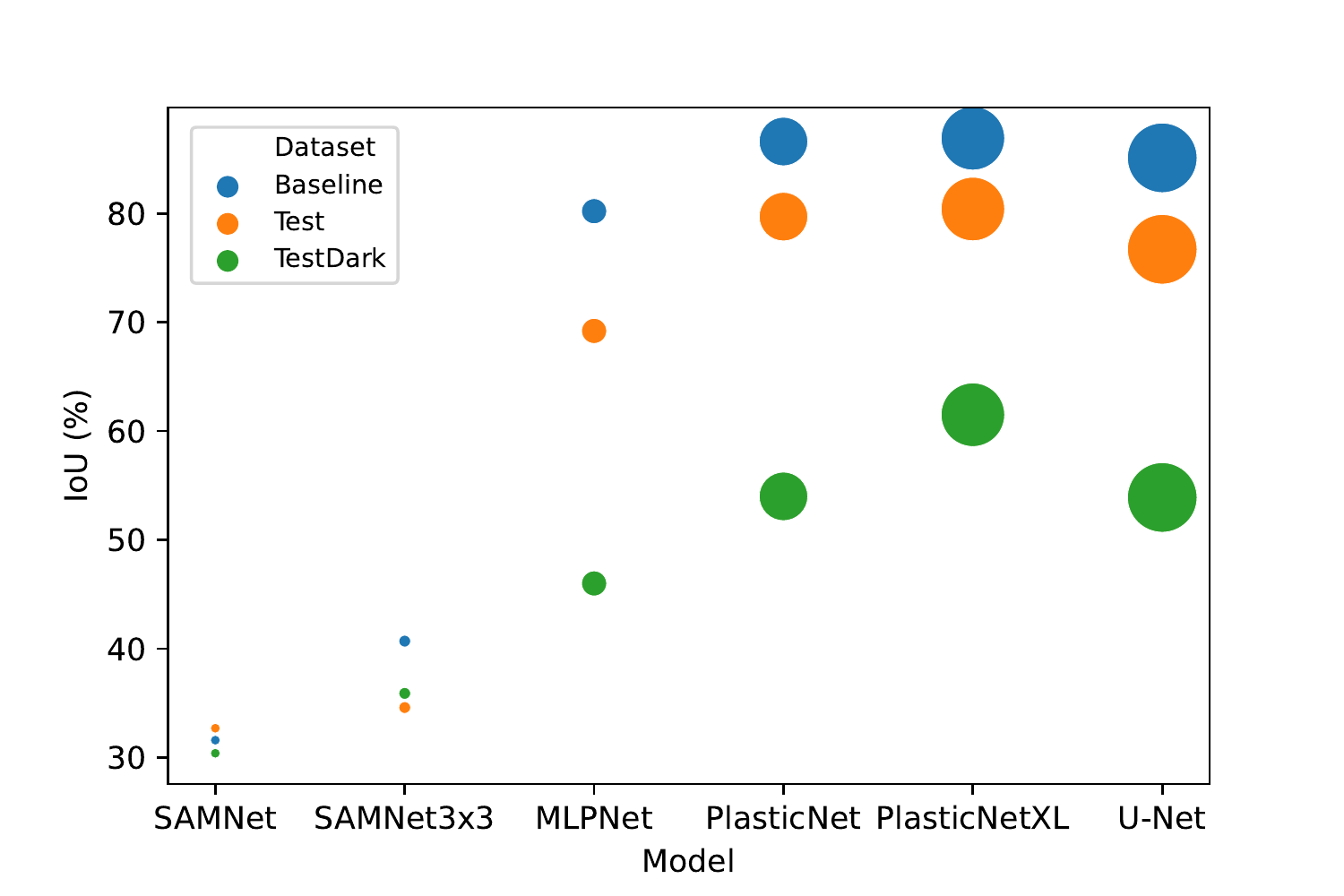} }}%
\caption{IoU of the best performing pre-processing method for each model. \\The size indicates the computational complexity of the model and the color indicates the testing set used.}
\label{fig:cpl}
\end{figure}

Figure~\ref{fig:cpl} shows a graph that relates the best IoU for all the models to the computational complexity for that model and the testing subset used. As can be seen, wider and deeper models keep offering improvement in performance over all testing datasets, while U-net does not follow the same improvement pattern. Another interesting observation is that, while the baseline performance does not differ much between the two PlasticNets, the PlasticNetXL model is better able to handle the Test-Dark dataset and thus it is safe to say that it generalizes better. The remainder of this section contains more detailed analysis of the results.

\begin{table*}
\centering
\begin{tabular}{ll|rrr|rrr|rrr}
\toprule
 & Dataset & \multicolumn{3}{c}{Baseline-Test}   &  \multicolumn{3}{c}{Test}   &  \multicolumn{3}{c}{Test-Dark}\\
 & Pre-process  &   IoU &  Precision &  Recall &  IoU &  Precision &  Recall &  IoU &  Precision &  Recall \\
Model &  &       &            &         &       &             &          &       &             &          \\
\midrule
SAMNet & SpectralNorm &  \textbf{31.6} &       39.7 &    63.5 &  \textbf{32.7} &        41.3 &     51.4 &  22.3 &        29.0 &     49.4 \\
              & HyperHSV &  28.8 &       39.9 &    63.3 &  22.8 &        36.4 &     47.5 &  \textbf{30.4} &        47.8 &     44.2 \\
\midrule              
SAMNet3x3 & SpectralNorm &  37.1 &       48.2 &    64.8 &  \textbf{34.6} &        47.7 &     50.8 &  25.2 &        32.4 &     54.5 \\
              & HyperHSV &  \textbf{40.7} &       54.8 &    69.1 &  32.1 &        50.6 &     50.8 &  \textbf{35.9} &        59.8 &     45.8 \\
\midrule              
MLPNet & None &  74.0 &       85.7 &    83.6 &  62.0 &        82.4 &     70.9 &  25.0 &        30.6 &     60.6 \\
              & FirstDeriv &  \textbf{80.2} &       95.0 &    83.5 &  61.3 &        95.9 &     63.1 &  39.3 &        48.6 &     56.1 \\
              & SecondDeriv &  72.8 &       94.6 &    75.6 &  54.9 &        97.5 &     56.0 &  27.4 &        32.0 &     50.4 \\
              & HyperHSV &  67.2 &       87.8 &    72.7 &  56.8 &        86.4 &     61.4 &  43.1 &        85.2 &     45.8 \\
              & LogDeriv &  36.1 &       52.0 &    41.6 &  30.2 &        43.8 &     36.9 &  21.7 &        28.2 &     24.0 \\
              & SpectralNorm &  79.0 &       91.8 &    84.6 &  \textbf{69.2} &        88.2 &     75.8 &  \textbf{46.0} &        76.6 &     51.3 \\
              & HyperHue &  64.5 &       81.3 &    73.0 &  52.6 &        71.1 &     62.8 &  32.3 &        40.6 &     46.6 \\
\cmidrule{2-11}     
              & \emph{mean} &  \emph{67.7} &       \emph{84.0} &    \emph{73.5} &  \emph{55.3} &        \emph{80.8} &     \emph{61.0} &  \emph{33.5} &        \emph{48.8} &     \emph{47.8} \\
\midrule              
U-Net & None &  84.5 &       90.6 &    92.4 &  65.6 &        81.1 &     74.2 &  \textbf{53.9} &        84.8 &     61.5 \\
              & FirstDeriv &  \textbf{85.1} &       95.0 &    89.0 &  65.7 &        81.4 &     81.4 &  51.6 &        93.0 &     54.6 \\
              & SecondDeriv &  82.7 &       92.2 &    88.7 &  47.5 &        60.5 &     74.7 &  42.7 &        86.1 &     46.5 \\
              & HyperHSV &  83.7 &       89.8 &    92.3 &  74.8 &        86.1 &     83.5 &  51.1 &        81.8 &     61.0 \\
              & LogDeriv &  66.4 &       81.9 &    76.4 &  31.7 &        39.5 &     40.5 &  23.0 &        70.0 &     27.0 \\
              & SpectralNorm &  78.2 &       92.5 &    83.4 &  69.1 &        89.9 &     74.8 &  40.4 &        78.1 &     47.9 \\
              & HyperHue &  84.1 &       90.2 &    92.3 &  \textbf{76.7} &        86.0 &     86.7 &  52.1 &        81.6 &     61.2 \\
\cmidrule{2-11}     
              & \emph{mean} & \textcolor{blue}{\emph{80.7}} &       \emph{90.3} &    \emph{87.8} &  \textcolor{blue}{\emph{61.6}} &        \emph{74.9} &     \emph{73.7} &  \textcolor{blue}{\emph{45.0}} &        \emph{82.2} &     \emph{51.4} \\
\midrule
PlasticNet & None &  28.7 &       63.0 &    38.8 &  25.6 &        49.8 &     36.0 &  23.8 &        52.1 &     34.2 \\
              & FirstDeriv &  86.5 &       94.1 &    91.2 &  77.8 &        93.6 &     81.8 &  37.8 &        77.8 &     46.0 \\
              & SecondDeriv &  80.3 &       94.3 &    84.1 &  64.9 &        96.8 &     66.4 &  30.6 &        98.4 &     31.5 \\
              & HyperHSV &  86.2 &       93.6 &    91.4 &  77.1 &        94.2 &     80.7 &  42.9 &        69.3 &     54.3 \\
              & LogDeriv &  56.9 &       81.0 &    64.0 &  38.4 &        62.3 &     45.6 &  22.6 &        48.3 &     27.6 \\
              & SpectralNorm &  \textbf{86.6} &       93.7 &    91.8 &  \textbf{79.7} &        92.8 &     85.3 &  52.3 &        84.8 &     59.4 \\
              & HyperHue &  85.3 &       91.2 &    92.7 &  65.9 &        74.0 &     84.8 &  \textbf{54.0} &        79.7 &     63.3 \\
\cmidrule{2-11}     
              & \emph{mean} &  \emph{72.9} &       \emph{87.3} &    \emph{79.1} &  \emph{61.3} &        \emph{80.5} &     \emph{68.7} &  \emph{37.7} &        \emph{72.9} &     \emph{45.2} \\
\midrule              
PlasticNetXL & None &  39.8 &       71.1 &    48.6 &  33.8 &        71.0 &     47.0 &  31.4 &        57.1 &     44.8 \\
              & FirstDeriv &  86.4 &       93.9 &    91.4 &  75.3 &        90.0 &     81.2 &  36.5 &        80.4 &     45.9 \\
              & SecondDeriv &  60.5 &       91.6 &    64.5 &  36.1 &        78.9 &     42.2 &  21.9 &        95.2 &     23.0 \\
              & HyperHSV &  \textbf{86.9} &       93.8 &    91.9 &  77.0 &        94.8 &     80.4 &  52.1 &        87.3 &     56.0 \\
              & LogDeriv &  67.6 &       86.8 &    74.4 &  35.3 &        59.9 &     39.8 &  20.1 &        39.2 &     29.2 \\
              & SpectralNorm &  86.1 &       93.3 &    91.7 &  \textbf{80.4} &        91.7 &     87.1 &  \textbf{61.5} &        91.9 &     64.6 \\
              & HyperHue &  86.6 &       93.5 &    91.9 &  71.0 &        79.6 &     87.3 &  56.2 &        88.0 &     60.6 \\
\cmidrule{2-11}     
              & \emph{mean} &  \emph{73.4} &       \emph{89.1} &    \emph{79.2} &  \emph{58.4} &        \emph{80.8} &     \emph{66.4} &  \emph{40.0} &        \emph{77.0} &     \emph{46.3} \\
\bottomrule
\end{tabular}
\caption{\normalsize Results of all models, all pre-processing methods on all datasets, PlasticNetXL shows the best IoU on all datasets.}
\label{tab:results}
\end{table*}

\subsection{Models and pre-processing}
The PlasticNetXL model shows the overall best performance on all datasets. On the Baseline-Test subset this model achieves 86.9\% IoU, which is only slightly higher then the runner-up, which is 86.6\% for PlasticNet. This is interesting given the fact that PlasticNet has roughly half the computational complexity of PlasticNetXL. The IoU of U-Net is 1.5\% lower (85.1\%) which means that the added architectural complexity not only does not play favorably for segmenting spectral pixels, but also adds additional computational complexity. This seems to confirm the hypothesis that architectures that are more tailored towards the specific application (e.g. the PlasticNets) outperform architectures developed for more general purposes like U-Net.

On the Test subset, the best IoU for each model shows a similar pattern, here also both PlasticNets perform best (80.4\% and 79.7\%), which is achieved with SpectralNorm pre-processing in both cases.

Pre-processing the data with either SpectralNorm or HyperHue gives the best performance for all models on the Test dataset. For most models this performance gain is substantial, for example the gain for MLPNet is 69.2\% - 61.0\% = 7.2\%. This seems to confirm the hypothesis that pre-processing algorithms that normalize intensity play an important role in creating models that generalize better.

The pre-processing methods that are based on derivatives almost never achieve the best results. The only case in which this happens is for FirstDeriv with U-Net on the Baseline-Test subset, however this effect vanished when using more realistic testing data.

\subsection{Depth and width of the models}
SAMNet is a model with a shallow architecture that does not take spatial context into account and has a footprint of $1 \times 1$. Its performance can be considered as the baseline performance for linear models on this dataset. The simple trick of extending the convolution window of SAMNet to $3 \times 3$ achieves a 9.1\%, 1.9\% and 5.5\% added performance on the Baseline-Test, Test and Test-Dark subsets, respectively. This evidently shows that the averaging effect of using a larger filter seems to be a good prior for this type of HS data. 

When using a somewhat deeper HS model like an MLPNet, an added performance of 39.5\%, 35\% and 10\% on the Baseline-Test, Test and Test-Dark subsets, respectively, is achieved. This shows that using a deeper non-linear model can almost double the performance (from 40.7\% to 80.2\% on the Baseline-Test subset) for this type of data.

The combination of deep models and wide footprints show the best performance indicated by the PlasticNets. Other common deep-design patterns like max-pooling layers and horizontal connections seem to be mostly detrimental (as indicated by the U-Net performance). However, averaged over all pre-processing methods, U-Net shows the best performance on all subsets, 80.7\%, 61.6\% and 45.0\%, respectively. This shows that the more sophisticated architecture of U-Net seems to be better able to handle the disturbances of the input signal caused by pre-processing, even though they might be detrimental for other architectures.

\subsection{Low exposure evaluation}
When using a Test-Dark subset, for which the distribution of the testing data is largely outside of the original training distribution, the performance degrades considerably for all methods (the PlasticNet model shows a more than 25\% decrease in IoU). Also, the results become very noisy. Interestingly, the most simple models (i.e. the SAMNets) show stable performance (or even a slightly increased performance between the Test and the Test-Dark subsets). This shows that simplicity of linear models allows for a more stable segmentation performance, but unfortunately the absolute performance is low. Moreover, the pre-processing methods seems to not be able to keep the performance at an acceptable level. It can even be seen that for U-Net using no pre-processing seems to achieve the best performance on the Test-Dark subset. This could be caused by the fact that the signal Test-Dark subset is already of low quality caused by the low-exposure, and the pre-processing methods tend to degrade them even further. Hence, not degrading the signal with pre-processing leads to the best IoU for U-Net.

\subsection{Polymer classification analysis}
Heretofore, the average performance of the model in combination with pre-processing methods has been discussed. In this subsection, we extend this analysis for the best-performing pre-processing and model combinations on each of the testing subsets, for each polymer type class. The per-class IoU, precision and recall values are shown in Table~\ref{tab:classperf}. On the Baseline-Test subset the IoU of the best-performing polymer class is 89.0\% for PE, and the lowest IoU is measured for PET (79.5\%).

When analysing the performance on the Test subset, it degrades for some polymer types and improves for other polymer types, which is surprising because one would expect that the performance would decrease for all polymer types, given the fact that the flakes came from a different source than the training set. The PET segmentation performance increases from 79.5 \% to 81.8\%, while for PE, PS the performance decreases from 89.0\% to 68.0\% for PE and from 84.2\% to 71.1\% for PS. As already seen in the average IoU scores in the previous subsections, the performance on the Test-Dark subset further decreases for all polymer types.

These results do not show a clear discernible pattern in the performance among polymer types. Initially it was expected that some polymer types might be easier to distinguish compared to others because of inherent differences in the spectral patterns. A possible reason why this is not observed in the results is probably because the original source material for producing the flakes, for each of the testing subsets, were taken randomly from the waste stream. The flakes can vary in composition and are not pure, this probably causes noise in the per-class IoU of the respective classes between subsets.


\subsection{Qualitative results}

\begin{figure*}[b]
    \centering
    \subfloat[Input]{\includegraphics[width=0.14\linewidth]{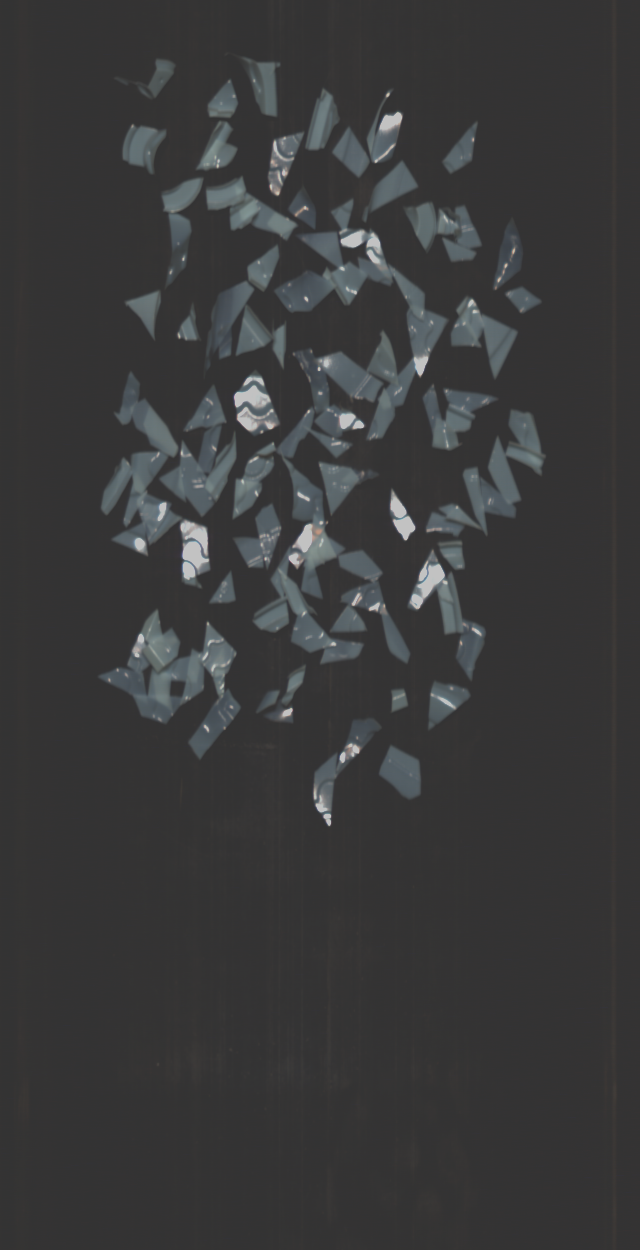}}
    \hspace{1mm}
    \subfloat[Target]{\includegraphics[width=0.14\linewidth]{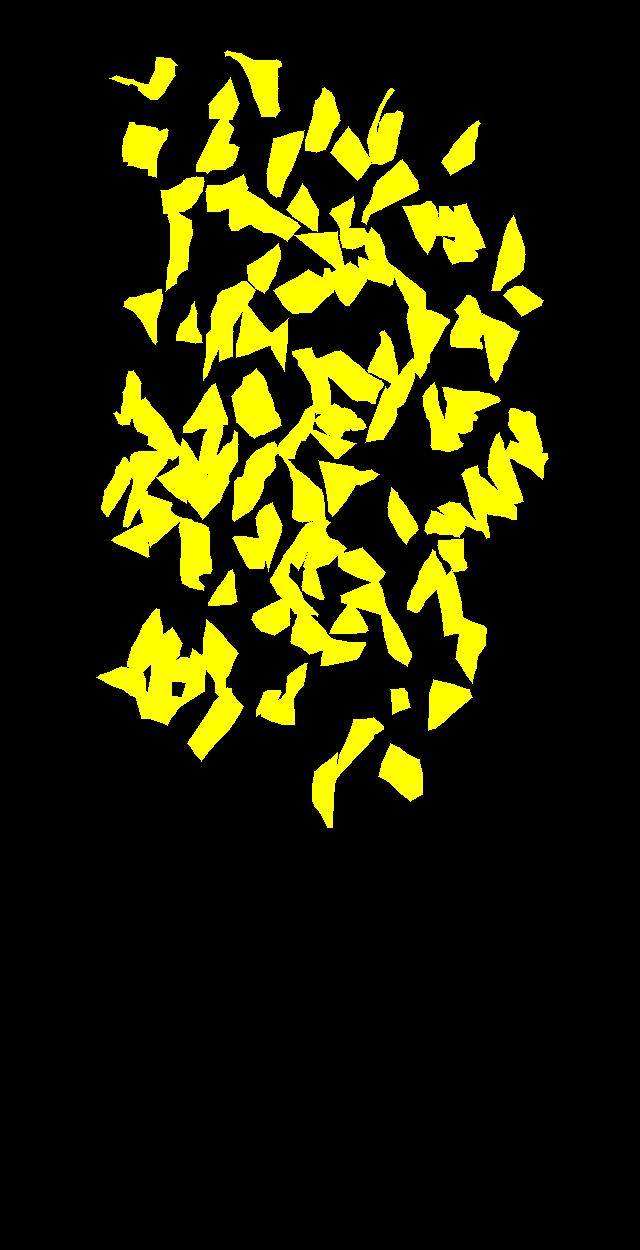}}
    \hspace{1mm} \\
    
    \subfloat[SAMNet]{\includegraphics[width=0.14\linewidth]{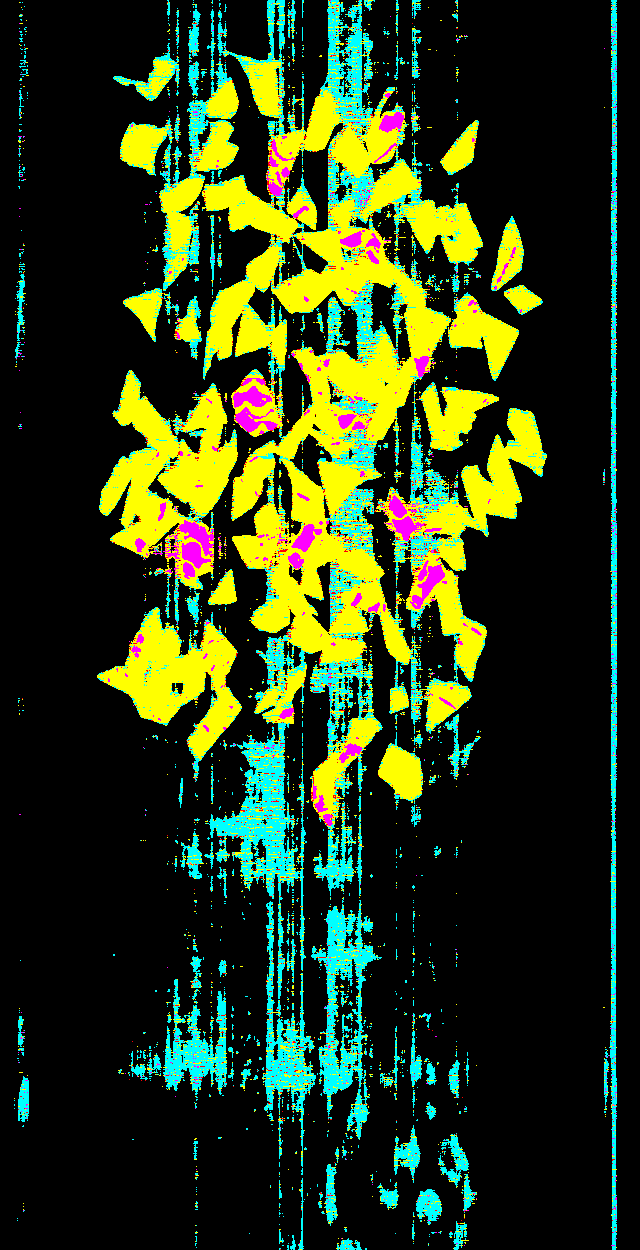}}
    \hspace{1mm}
    \subfloat[SAMNet3x3]{\includegraphics[width=0.14\linewidth]{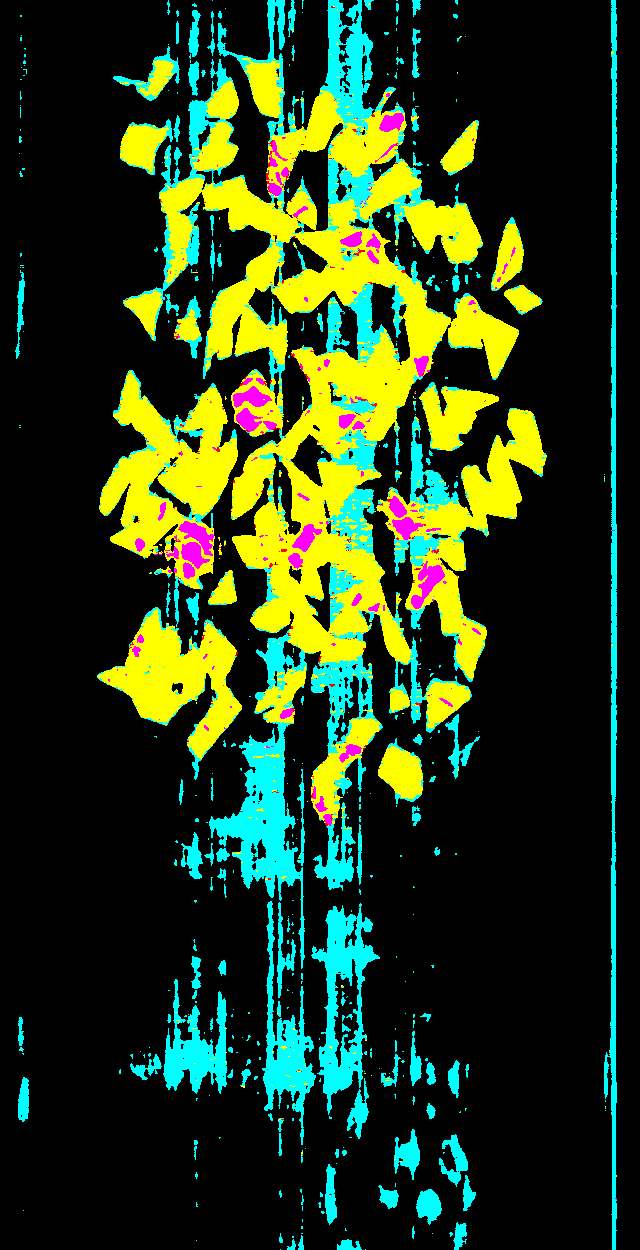}} \\

    \subfloat[None - PlasticNet]{\includegraphics[width=0.14\linewidth]{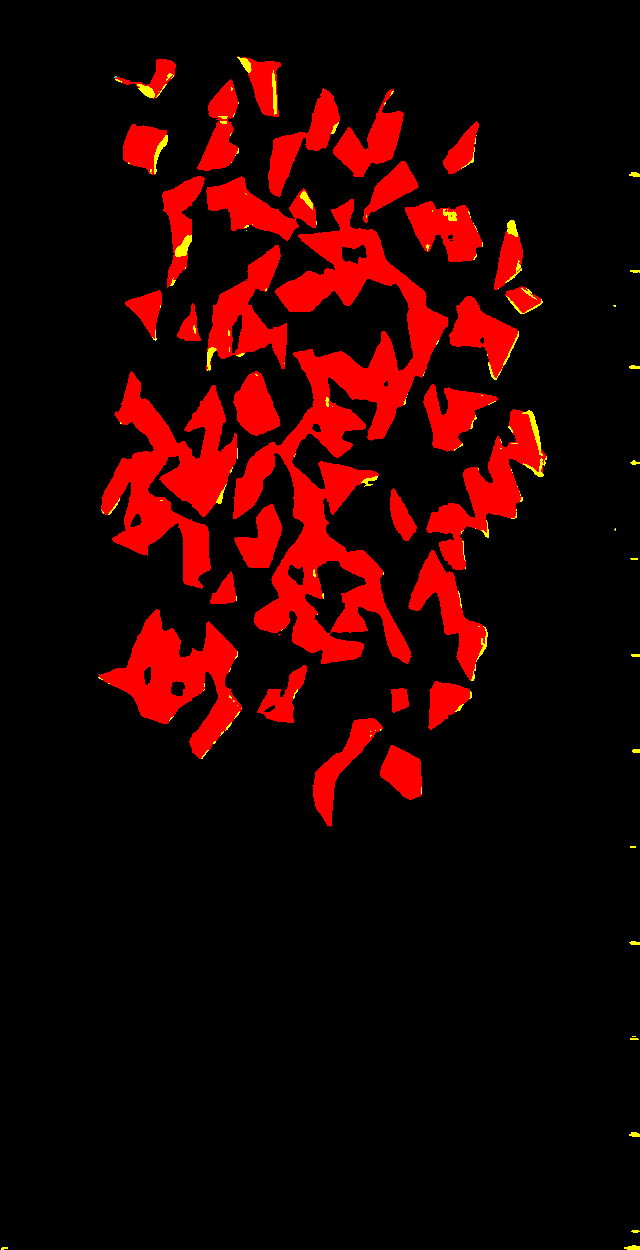}}
    \hspace{1mm}
    \subfloat[None - PlasticNetXL]{\includegraphics[width=0.14\linewidth]{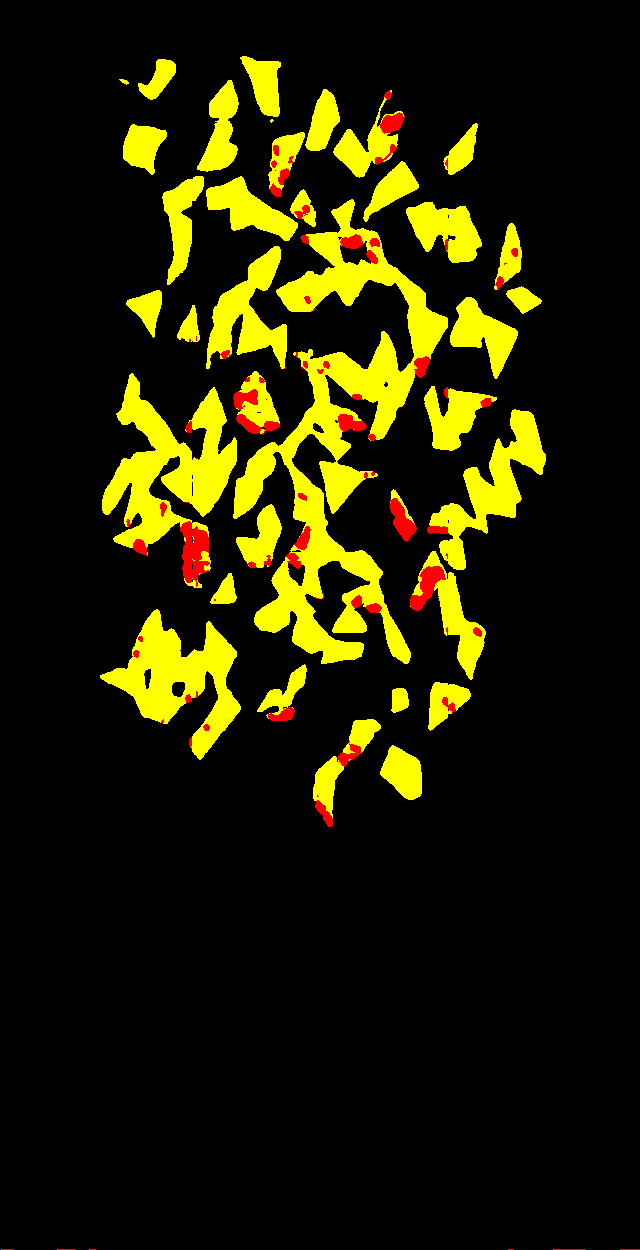}}
    \hspace{1mm}
    \subfloat[None - MLPNet]{\includegraphics[width=0.14\linewidth]{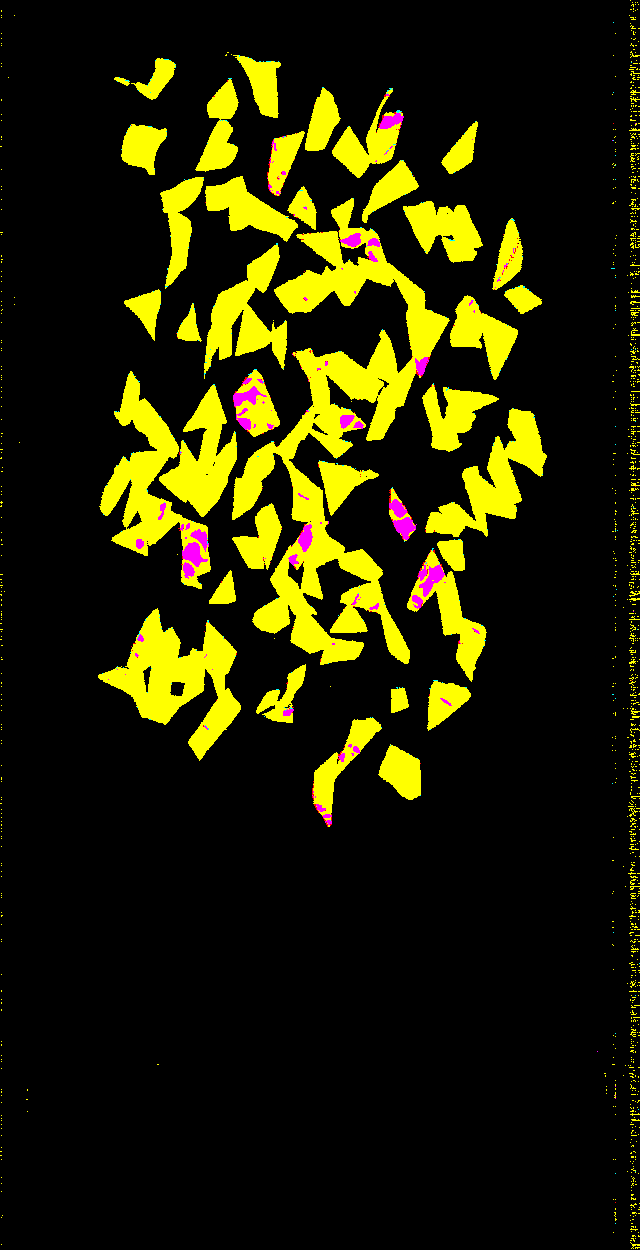}}
    \hspace{1mm}
    \subfloat[None - U-Net]{\includegraphics[width=0.14\linewidth]{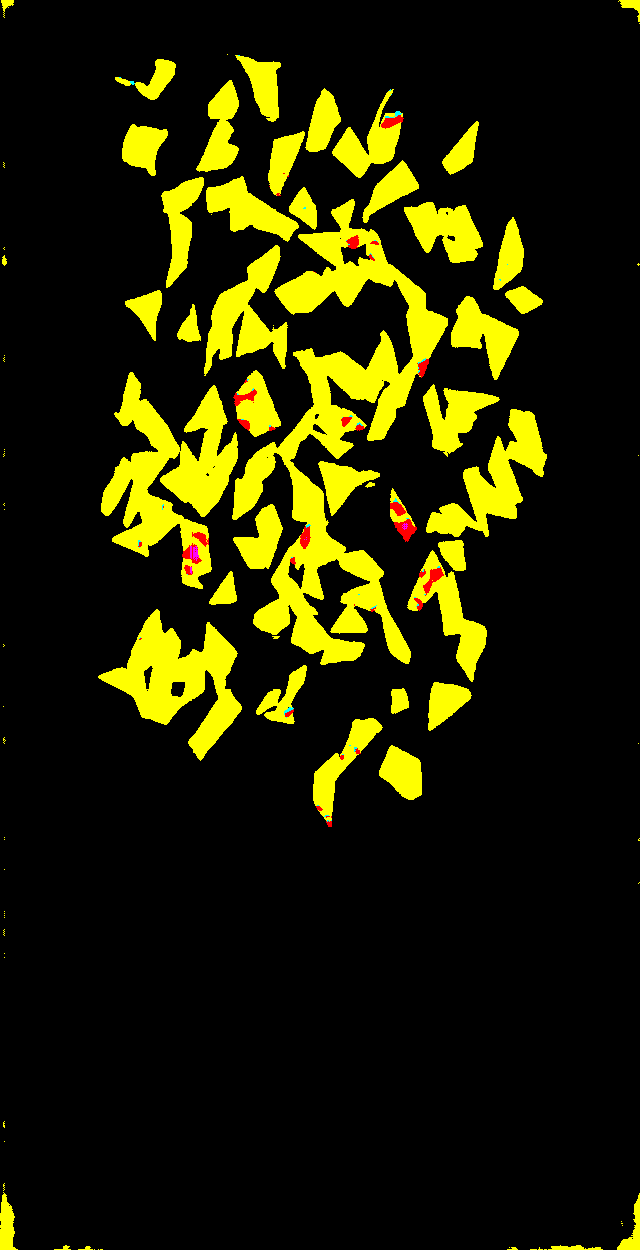}}
    \hspace{1mm}

    \subfloat[S.Norm - PlasticNet]{\includegraphics[width=0.14\linewidth]{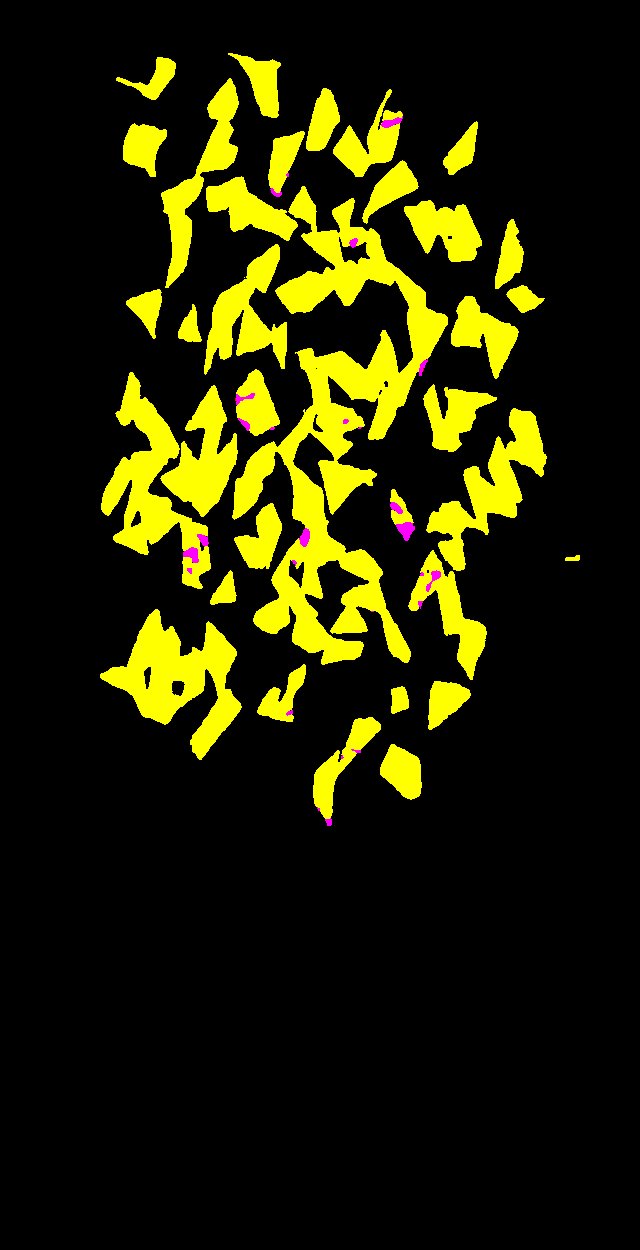}}
    \hspace{1mm}
    \subfloat[S.Norm - PlasticNetXL]{\includegraphics[width=0.14\linewidth]{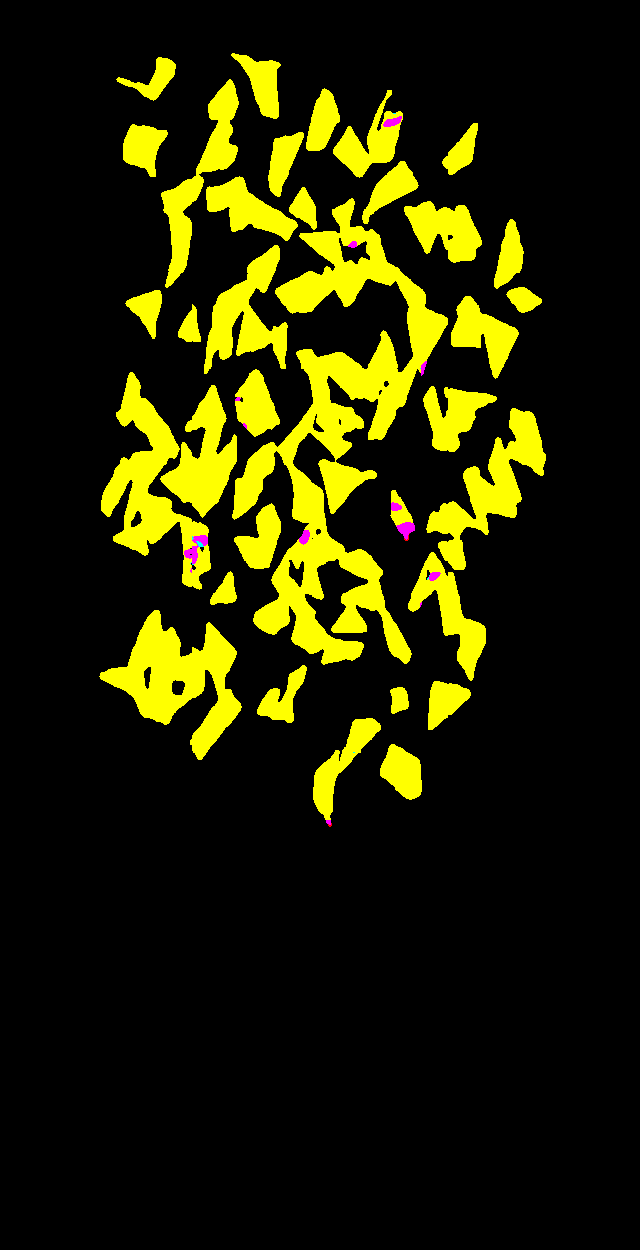}}
    \hspace{1mm}
    \subfloat[S.Norm - MLPNet]{\includegraphics[width=0.14\linewidth]{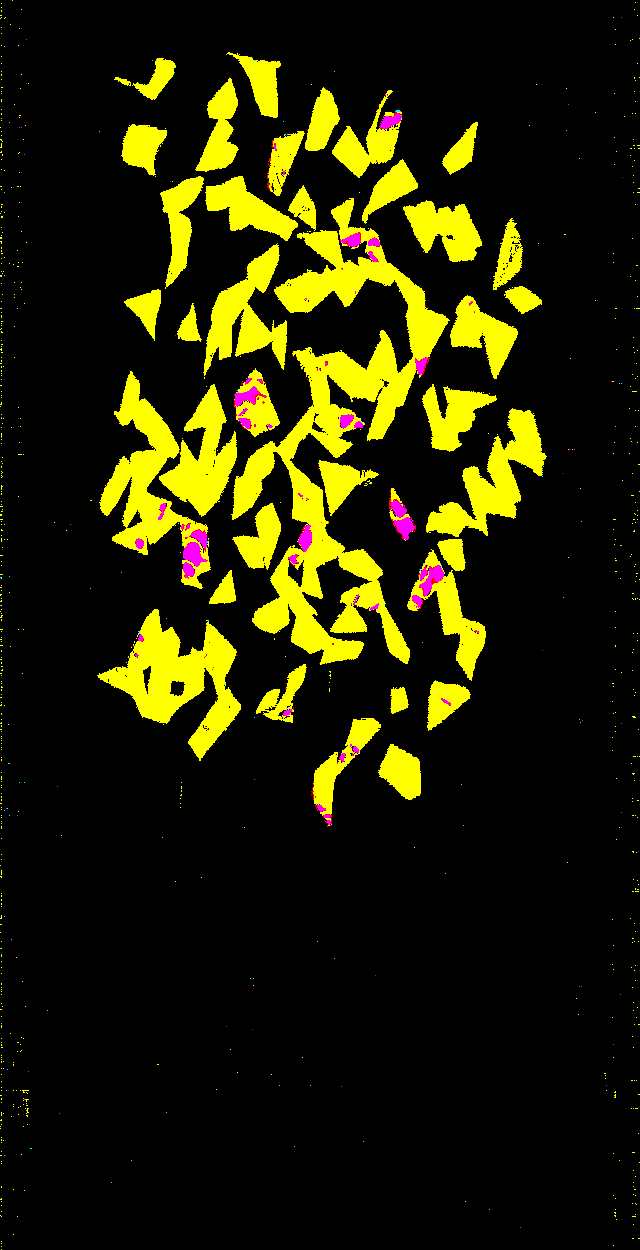}}
    \hspace{1mm}
    \subfloat[S.Norm - U-Net]{\includegraphics[width=0.14\linewidth]{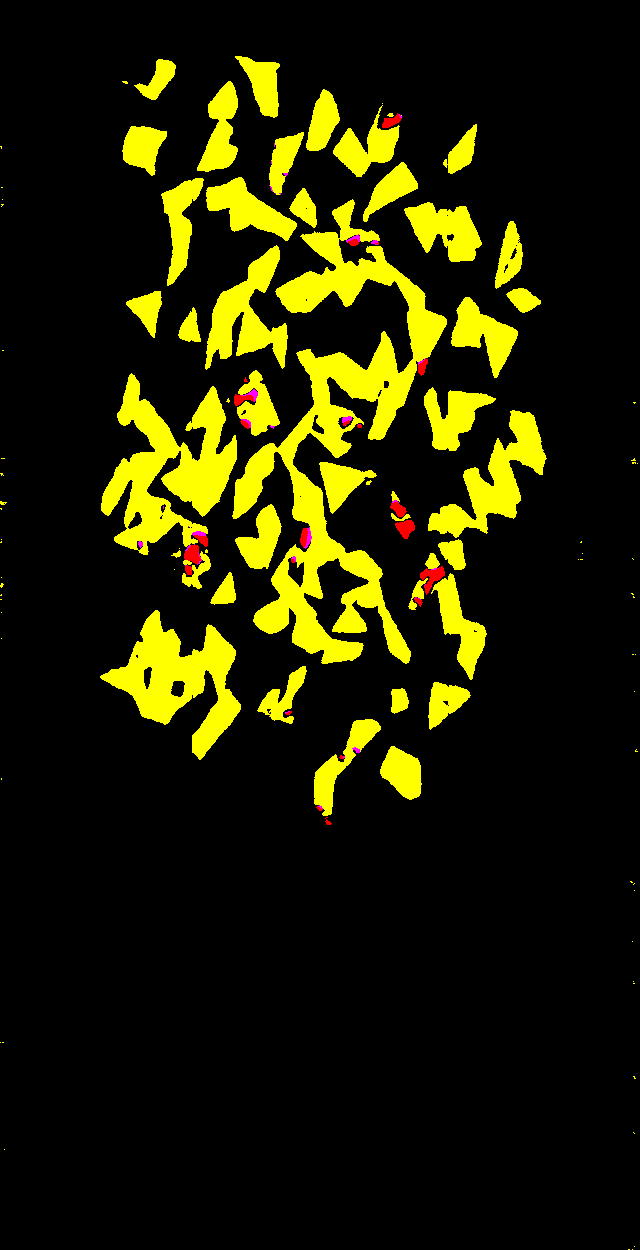}}
    \hspace{1mm}
\caption{Results for multiple PS flakes in the Baseline-Test dataset.}
\label{fig:baselineresults}
\end{figure*}

\begin{figure*}[b]
    \centering
    \subfloat[Input]{\includegraphics[width=0.14\linewidth]{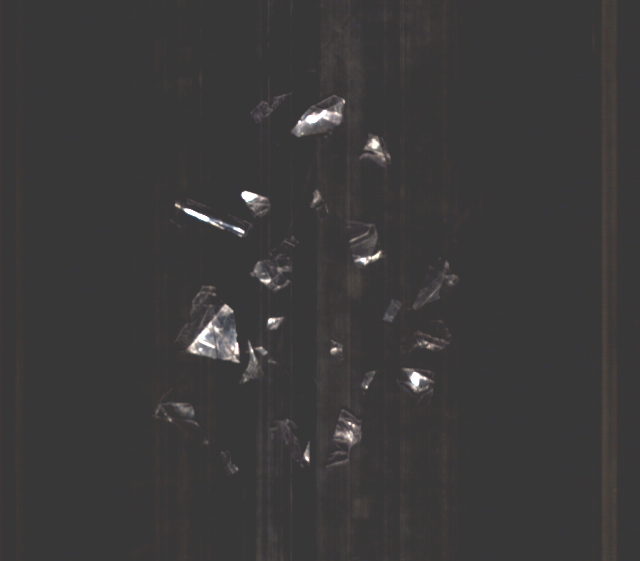}}
    \hspace{1mm}
    \subfloat[Target]{\includegraphics[width=0.14\linewidth]{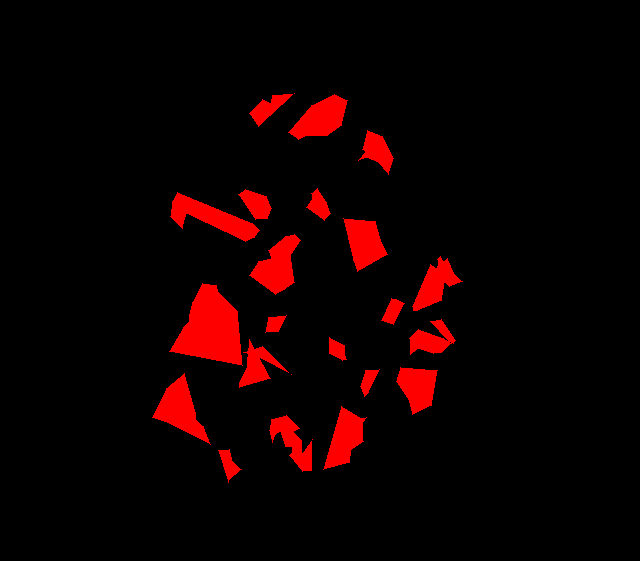}}
    \hspace{1mm} \\
    
    \subfloat[SAMNet]{\includegraphics[width=0.14\linewidth]{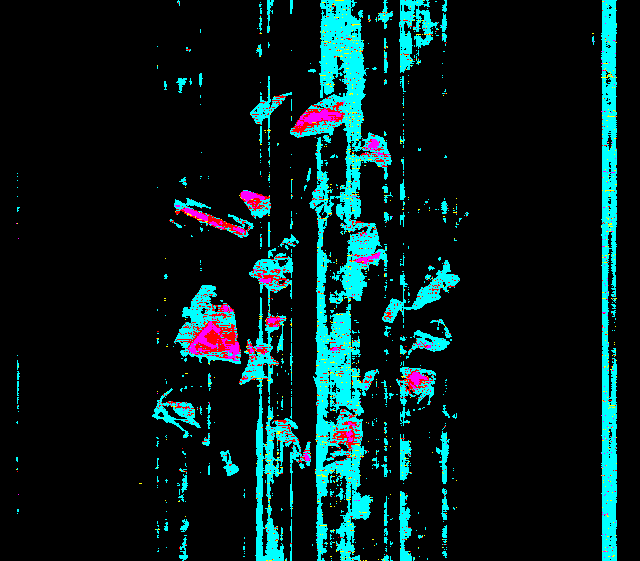}}
    \hspace{1mm}
    \subfloat[SAMNet3x3]{\includegraphics[width=0.14\linewidth]{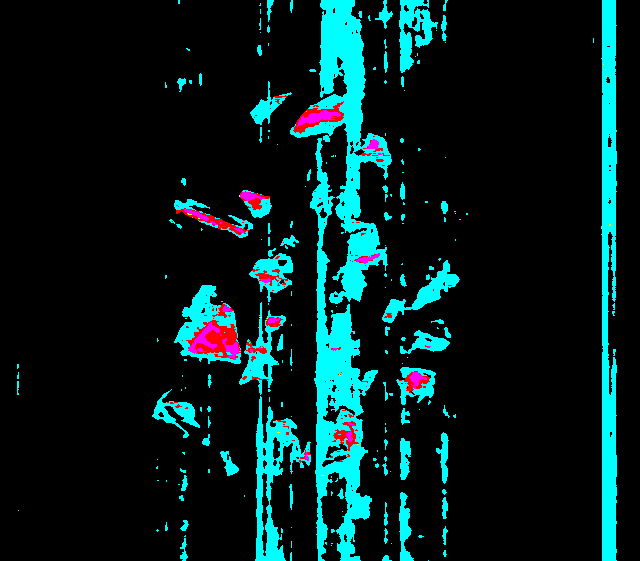}} \\

    \subfloat[None - PlasticNet]{\includegraphics[width=0.14\linewidth]{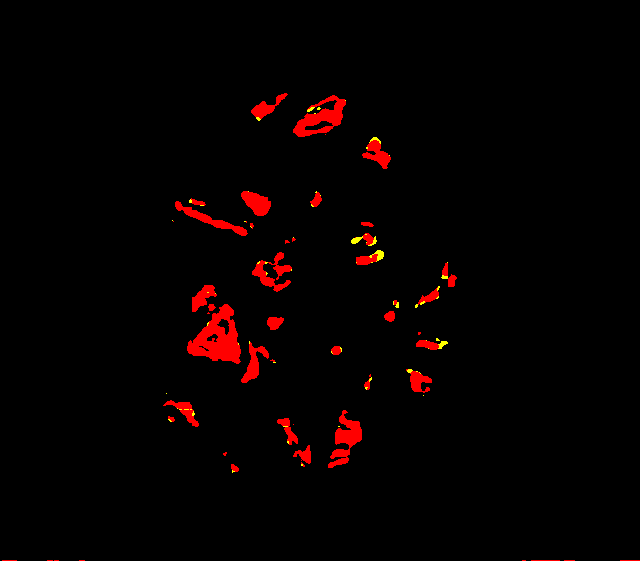}}
    \hspace{1mm}
    \subfloat[None - PlasticNetXL]{\includegraphics[width=0.14\linewidth]{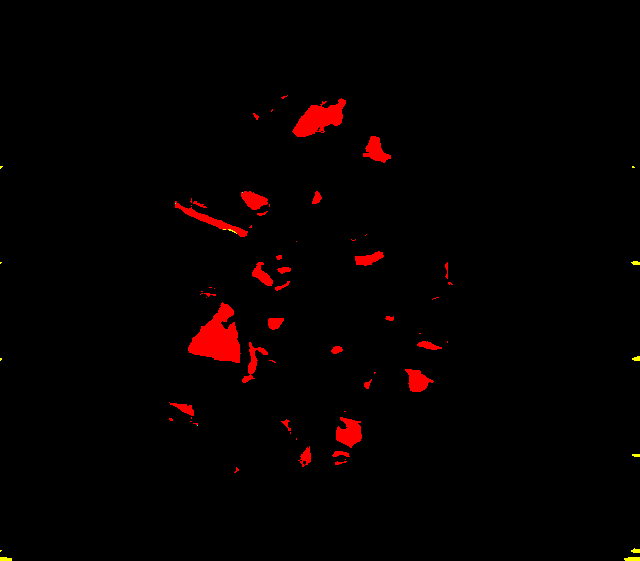}}
    \hspace{1mm}
    \subfloat[None - MLPNet]{\includegraphics[width=0.14\linewidth]{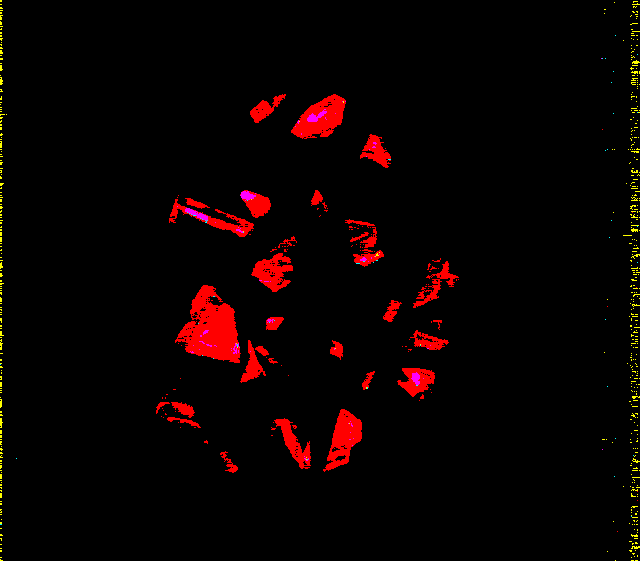}}
    \hspace{1mm}
    \subfloat[None - U-Net]{\includegraphics[width=0.14\linewidth]{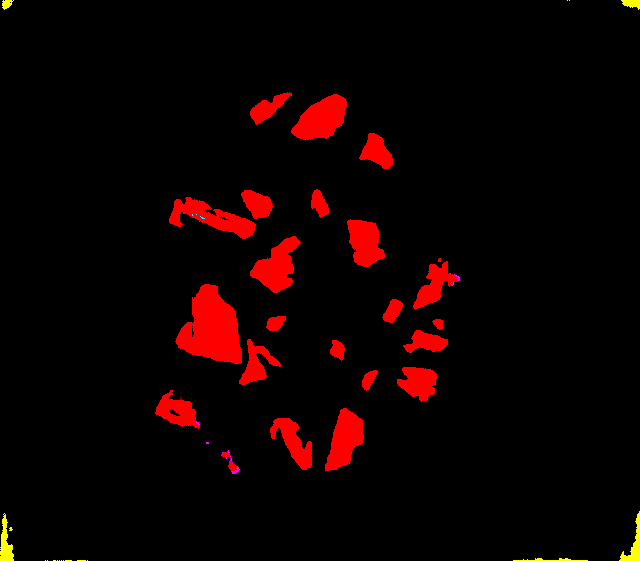}}
    \hspace{1mm}

    \subfloat[S.Norm - PlasticNet]{\includegraphics[width=0.14\linewidth]{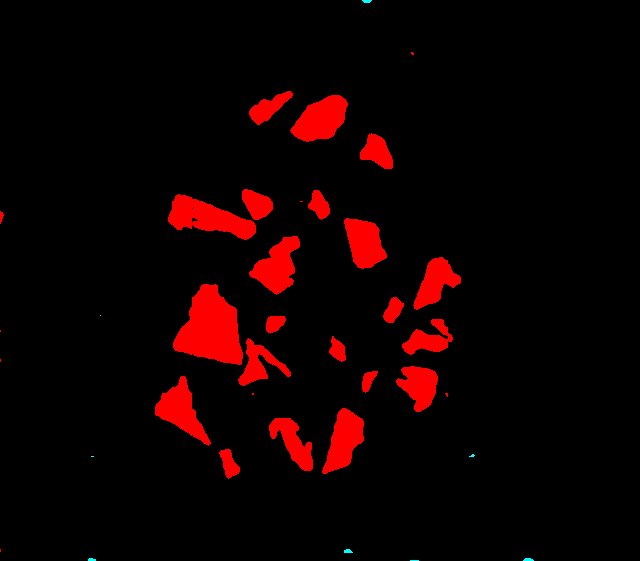}}
    \hspace{1mm}
    \subfloat[S.Norm - PlasticNetXL]{\includegraphics[width=0.14\linewidth]{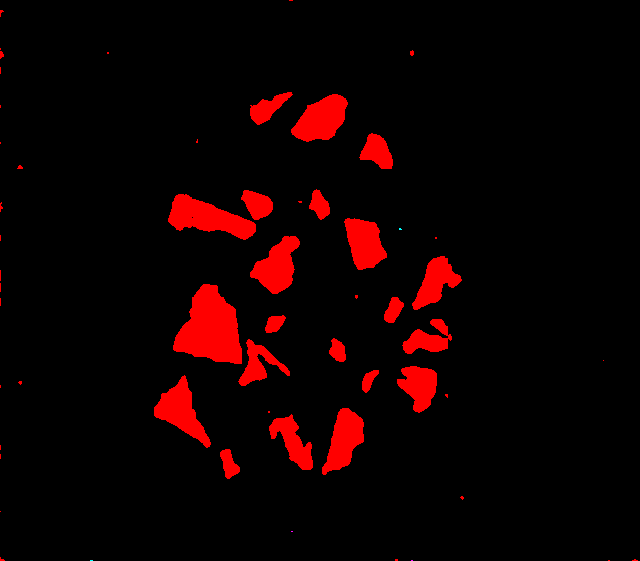}}
    \hspace{1mm}
    \subfloat[S.Norm - MLPNet]{\includegraphics[width=0.14\linewidth]{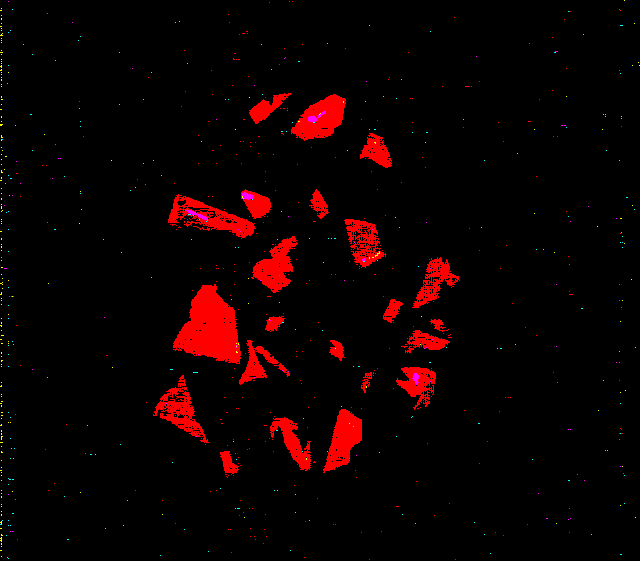}}
    \hspace{1mm}
    \subfloat[S.Norm - U-Net]{\includegraphics[width=0.14\linewidth]{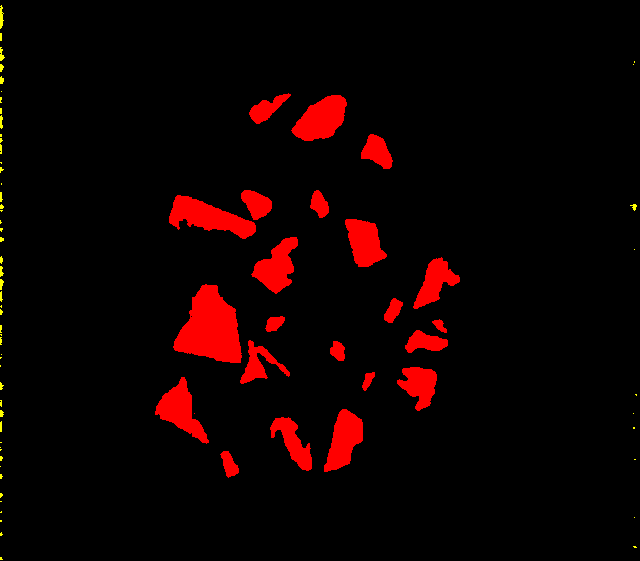}}
    \hspace{1mm}
\caption{Results for multiple PP flakes in the Test data set.}
\label{fig:testresults}
\end{figure*}

\begin{figure*}[b]
    \centering
    \subfloat[Input]{\includegraphics[width=0.14\linewidth]{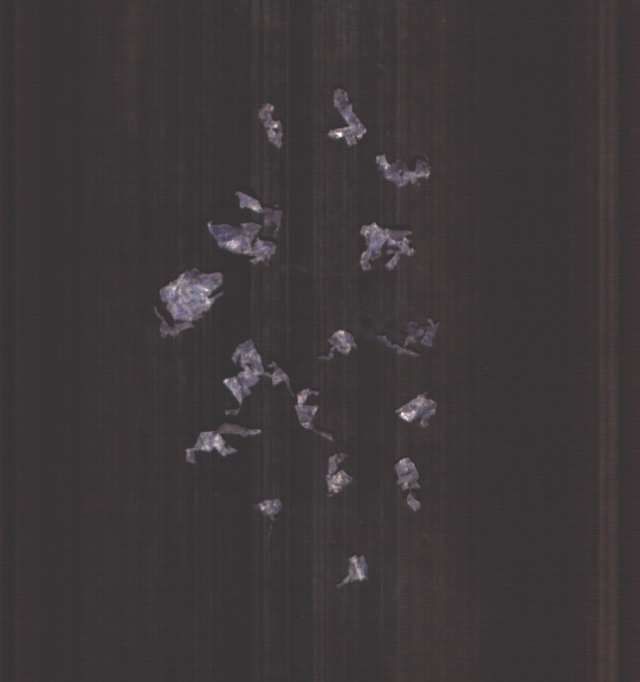}}
    \hspace{1mm}
    \subfloat[Target]{\includegraphics[width=0.14\linewidth]{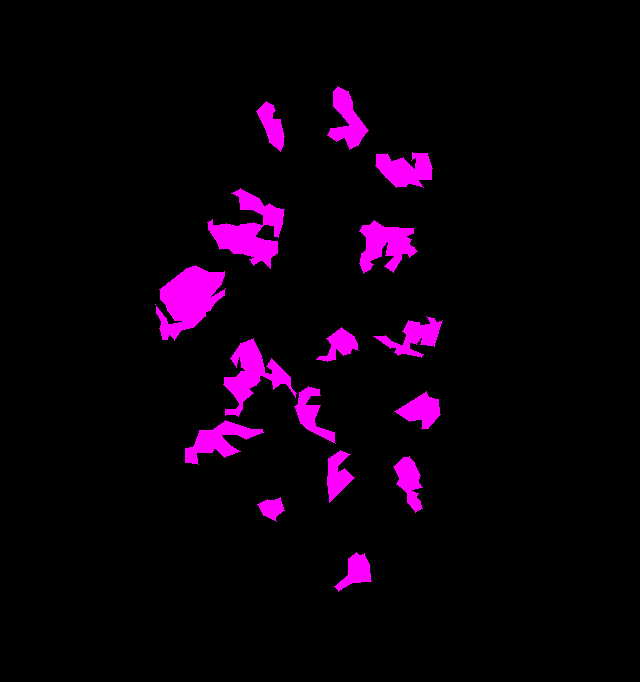}}
    \hspace{1mm} \\
    
    \subfloat[SAMNet]{\includegraphics[width=0.14\linewidth]{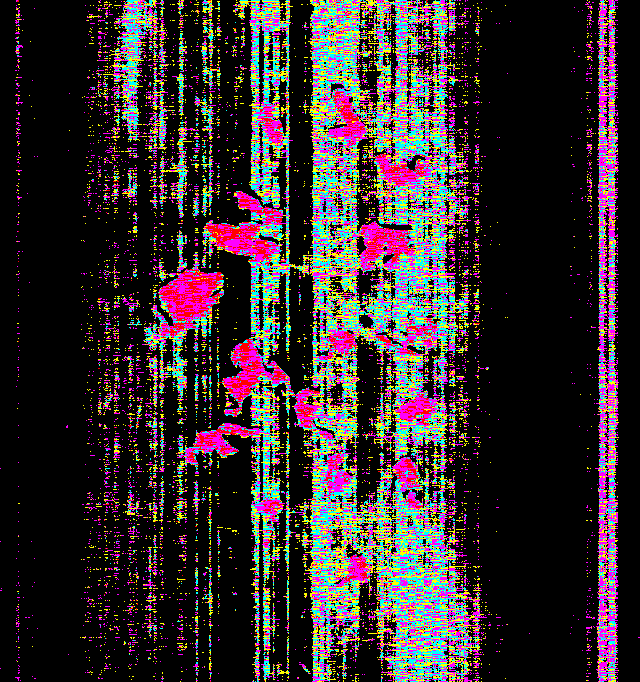}}
    \hspace{1mm}
    \subfloat[SAMNet3x3]{\includegraphics[width=0.14\linewidth]{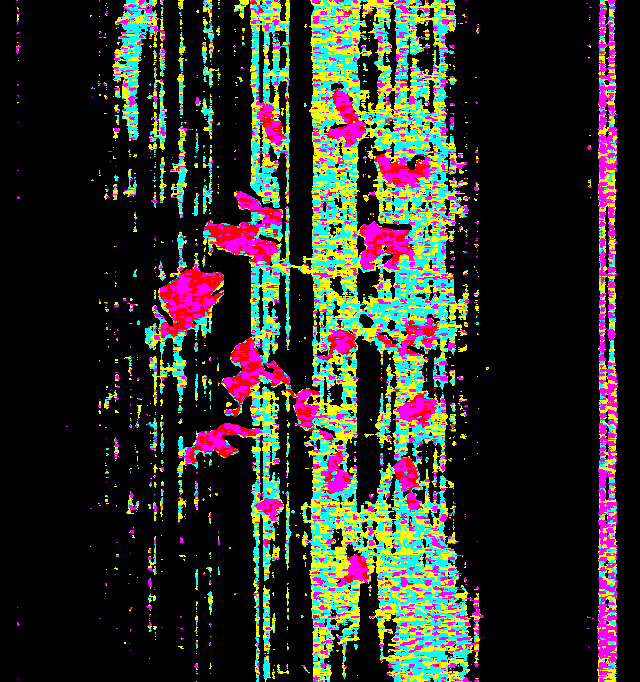}} \\

    \subfloat[None - PlasticNet]{\includegraphics[width=0.14\linewidth]{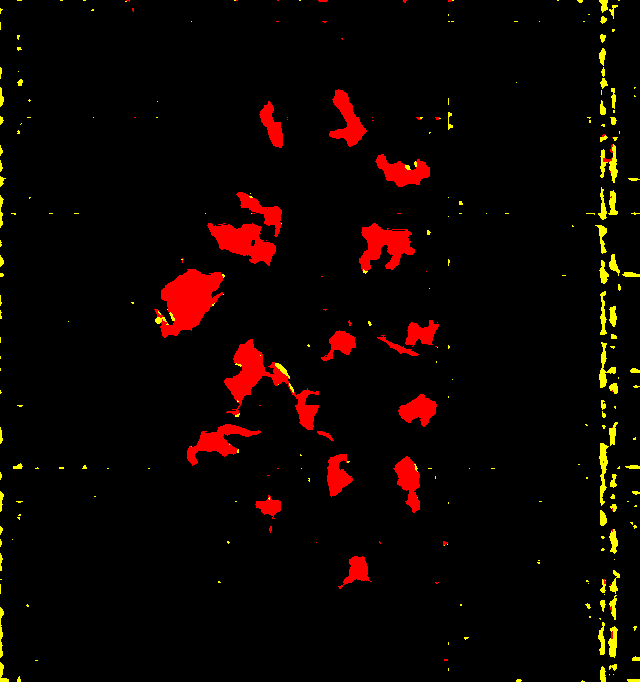}}
    \hspace{1mm}
    \subfloat[None - PlasticNetXL]{\includegraphics[width=0.14\linewidth]{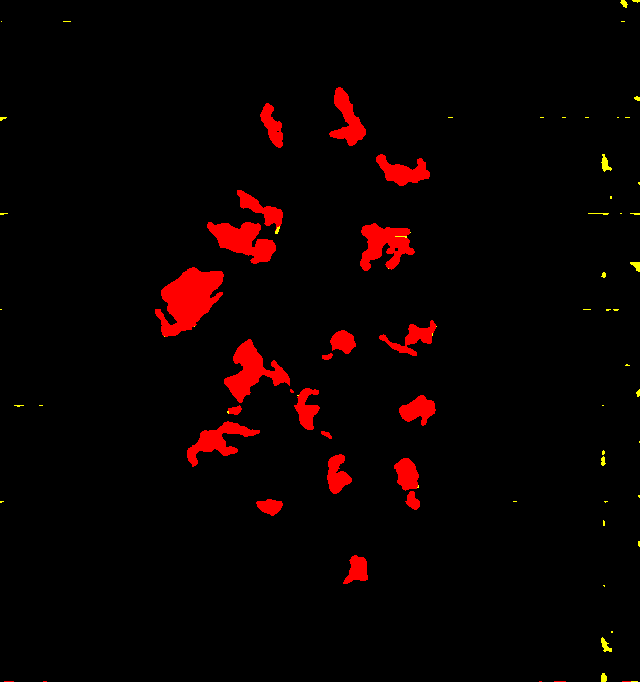}}
    \hspace{1mm}
    \subfloat[None - MLPNet]{\includegraphics[width=0.14\linewidth]{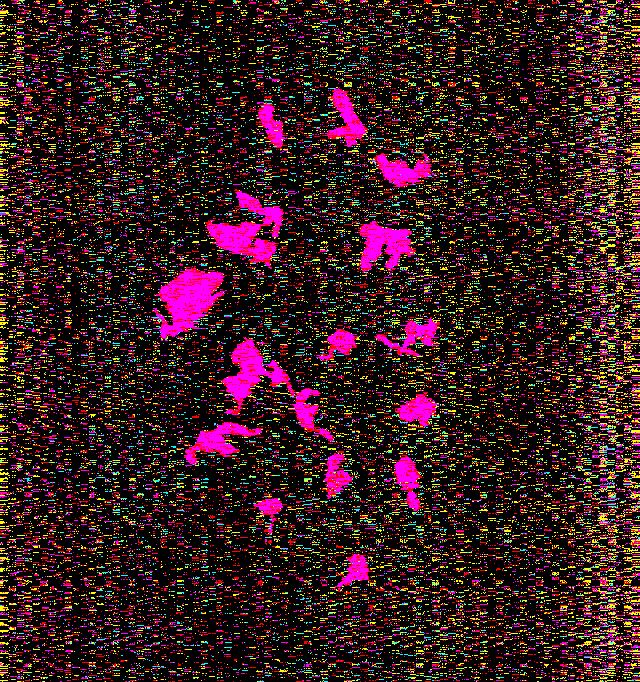}}
    \hspace{1mm}
    \subfloat[None - U-Net]{\includegraphics[width=0.14\linewidth]{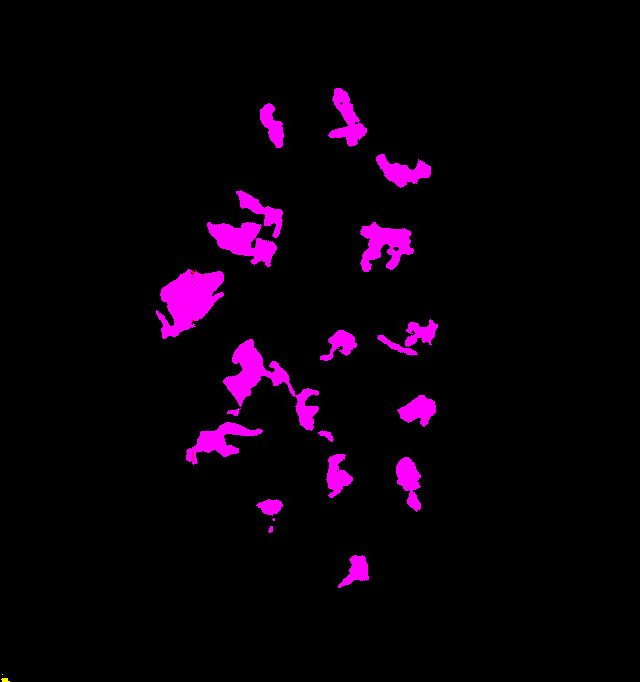}}
    \hspace{1mm}

    \subfloat[S.Norm - PlasticNet]{\includegraphics[width=0.14\linewidth]{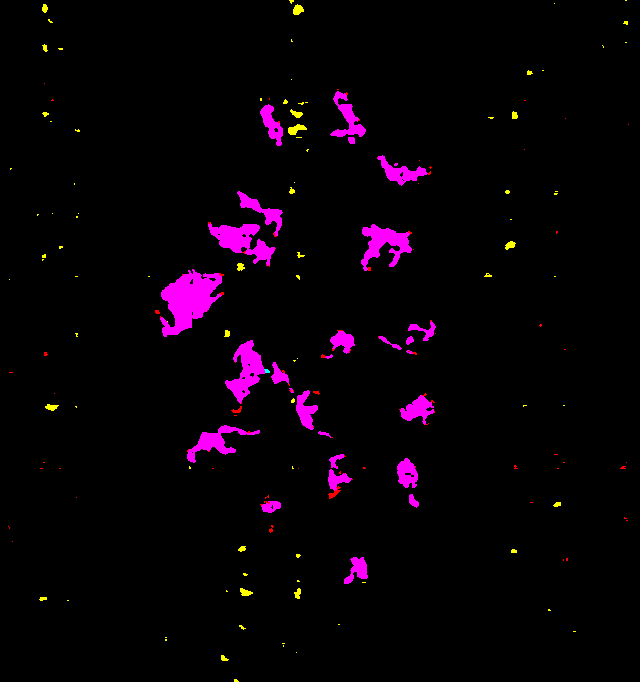}}
    \hspace{1mm}
    \subfloat[S.Norm - PlasticNetXL]{\includegraphics[width=0.14\linewidth]{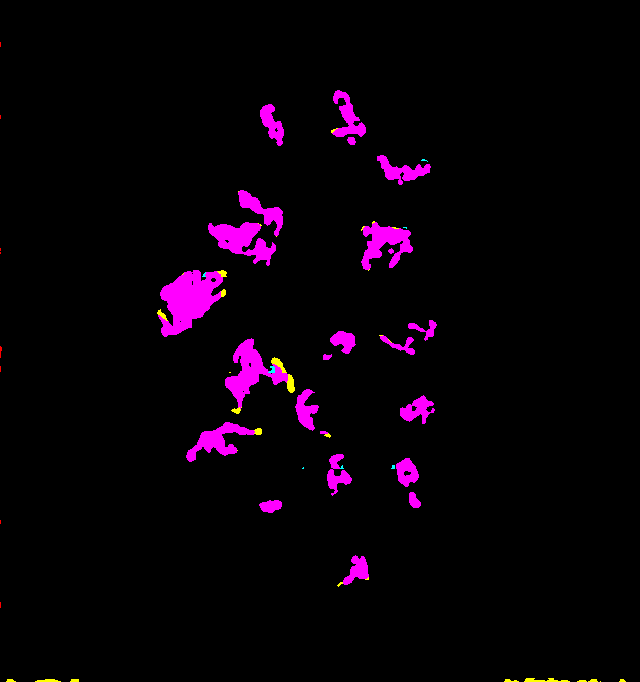}}
    \hspace{1mm}
    \subfloat[S.Norm - MLPNet]{\includegraphics[width=0.14\linewidth]{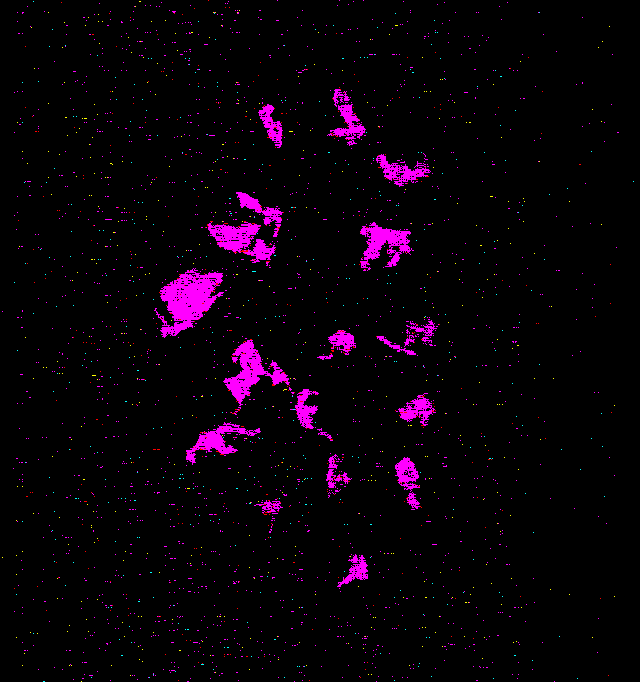}}
    \hspace{1mm}
    \subfloat[S.Norm - U-Net]{\includegraphics[width=0.14\linewidth]{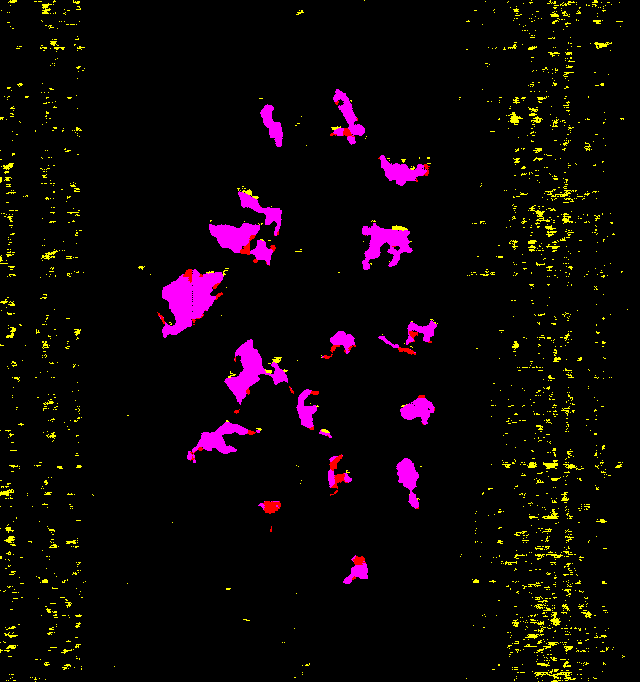}}
    \hspace{1mm}
\caption{Results for multiple PE flakes in the Test-Dark data set.}
\label{fig:darkresults}
\end{figure*}


The segmentation results for PS flakes of the Baseline-Test are shown in Figure~\ref{fig:baselineresults}, (a) being the HS cube mapped to RGB and (b) its corresponding target segmentation mask. It can be seen that both SAMNets show the poorest visual segmentation result, in (c) and (d). The noise is mainly in the background class, which is explained by the fact that the spectral normalization is built into the SAM algorithm, that causes background noise to be amplified especially if the norm of the spectral pixel is low (e.g. a dark background pixel). Already in the IoU results of the previous section it could be seen that the PlasticNets perform poorly when no pre-processing is used, which can clearly be explained by the fact that, for some images, PlasticNet misclassifies almost all pixels (e). Overall, when using SpectralNorm, the results for all trained models tend to be visually similar. 

When using the Test subset, there is a clear difference in performance between the usage of pre-processing methods (see Figure~\ref{fig:testresults}). Both PlasticNets and the MLPNet show a worse performance when using no pre-processing, particularly at the edges of the flakes. Also they look more noisy and some flakes are not segmented correctly at all, in Figure~\ref{fig:testresults} (e), (f) and (g). When visually inspecting these HS images of PP flakes, it seems like U-Net is better able to handle non-preprocessed images, as shown in Figure~\ref{fig:testresults} (h) and (i). This is consistent with the results from the Baseline-Test subset, which also the PlasticNets were less able to handle non pre-processed data. Finally, for this HS cube, the SAMNets are misclassifying all flakes, as shown in Figure~\ref{fig:testresults} (c) and (d).

To find the limits of the system the models have also been evaluated on the Test-Dark subset in Figure~\ref{fig:darkresults}. Here it can be seen that almost all models struggle with segmenting the flakes. Both PlasticNet models get the type of plastic entirely wrong when not using pre-processing, see images (e) and (f). Also there is much per-pixel noise visible in the results. 


\begin{table}
\centering
\begin{tabular}{llll}
\toprule
 &   IoU &  Precision &  Recall \\
Class      &       &            &         \\
\toprule
\multicolumn{4}{c}{\emph{HyperHSV + PlasticNetXL on the Baseline-Test dataset}}\\
\midrule
BG         &  98.7 &       99.3 &    99.4 \\
PE         &  89.0 &       94.7 &    93.7 \\
PET        &  79.5 &       91.9 &    85.5 \\
PP         &  83.0 &       91.7 &    89.7 \\
PS         &  84.2 &       91.5 &    91.3 \\
\toprule
\multicolumn{4}{c}{\emph{SpectralNorm + PlasticNetXL on the Test dataset}}\\
\midrule
BG         &  98.6 &       99.1 &    99.4 \\
PE         &  68.0 &       94.2 &    70.9 \\
PET        &  81.8 &       84.8 &    95.9 \\
PP         &  82.5 &       86.5 &    94.7 \\
PS         &  71.1 &       93.9 &    74.6 \\
\toprule
\multicolumn{4}{c}{\emph{SpectralNorm + PlasticNetXL on the Test-Dark dataset}}\\
\midrule
BG         &  97.4 &       97.7 &    99.7 \\
PE         &  44.4 &       97.3 &    44.9 \\
PET        &  52.7 &       95.4 &    54.1 \\
PP         &  66.6 &       95.4 &    68.8 \\
PS         &  46.3 &       73.6 &    55.5 \\
\bottomrule
\end{tabular}
\caption{\normalsize Per-class performance on flake types PE, PET, PP and PS using the best-performing models for each dataset.}
\label{tab:classperf}
\end{table}

\section{Conclusion}
In this paper, several models for segmenting polymer flakes in HS images were proposed. The developed models stem from multiple architectures: a CNN reformulations of a traditional linear model called SAMNet; a hyper-spectral and CNN reformulation of a traditional multi-layer perceptron called MLPNet; and two versions of PlasticNet, a model with a large footprint specifically designed for HS data. The models were compared with the popular U-Net model. Additionally, all models were trained on the same subset and each tested on three subsets with different properties, all containing pixel-level annotations for the flakes of four different polymer types: PE, PET, PS and PP.

The results of the experiments show that, our proposed model, PlasticNetXL achieves the best performance (max IoU is 86.9\%), however, on average, U-Net achieves the best results (mean IoU is 80.7\%). From this can be concluded that the model that was specifically designed for the task, PlasticNet, achieves the highest performance and is the preferred model for a practical application, thanks to its optimized complexity. At the same time, the more general U-Net architecture gives the best result on average, which means that it is more robust to various input signals. This observation could be regarded as an instance of the No Free Lunch (NFL) theorem \cite{ho2002simple} stating that a model that has the best performance on average is sub-optimal for a specific task.

Several pre-processing methods were also developed and tested using all the models. These methods could be divided into two categories: intensity normalizing methods and regular pre-processing methods. FirstDeriv, SecondDeriv and LogDeriv are pre-processing methods that put focus on the relative difference between spectral bands, HyperHSV and HyperHue are the hyper-spectral variants of color-space conversions that are common in machine vision, and SpectralNorm is a nomalization method inspired by the classic Spectral Angle Mapping (SAM) algorithm. 

The results of the experiments on pre-processing methods show that when training, validation and testing is performed on the same flakes and acquisition conditions, there is not a single favorite pre-preprocessing method that shows the best performance. However, when using HS images of flakes from different plastic source samples, there is a clear preference for pre-processing methods that normalize the input signal, like HyperHue and SpectralNorm. The derivative methods (first, second and logarithmic) generally do not perform well on this set. Furthermore, when performing inference on a dataset with a degraded signal (Test-Dark), the pre-processing methods are not helpful anymore. From this can be concluded that the generalizing behavior of models is improved by pre-processing, but this does not extend to situations for which the testing data is largely outside of the distribution of the original training data. 

A comparison of computational complexity for all the per-pixel segmentation models were given that show the trade-off between computational complexity and performance. It can be seen that the PlasticNets achieve a higher IoU with a lower computational complexity. The simplest non-linear models seem to be the most robust against variations in input signal like lowering the exposure of the camera, but at the same time do not perform very well. 

In the future, research could focus on expanding the dataset with a larger variation of polymer types and introduce more variety between polymers of the same type. This paper was mostly about per-pixel segmentation of HS images of polymer flakes. This approach is applicable for analyzing the content of flakes on a conveyor belt and could be expanded to investigate instance-level segmentation of polymer flakes. This might be helpful for calculating per-flake statistics such as the size distribution of flakes of the same type with the purpose of supporting the sorting process.

\ifCLASSOPTIONcompsoc
  \section*{Acknowledgments}
\else
  \section*{Acknowledgment}
\fi


\ifCLASSOPTIONcaptionsoff
  \newpage
\fi



\bibliographystyle{IEEEtran}
%
\bibliography{bibliography}

%

\begin{IEEEbiography}[{\includegraphics[width=1in,height=1in,clip,keepaspectratio]{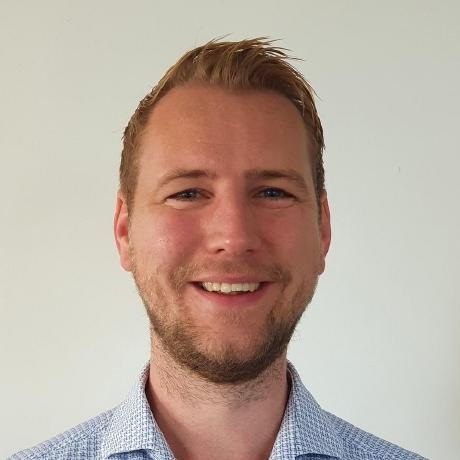}}]{Klaas~Dijkstra}
is~an~associate~professor~of applied sciences in Computer Vision and Data Science at the NHL Stenden University of Applied Sciences in Leeuwarden. His main research interests are in computer vision, deep learning and hyperspectral imaging. After completing his B.Eng. degree in technical information science in 2005, he has been active in the field of computer vision by doing applied research in several domains. In 2013 he obtained his M.Sc. degree from the Limerick Institute of Technology in Ireland, on the application of evolutionary algorithms and computer vision to the domain of microbiological analysis. In 2020 he obtained his PhD degree from the University of Groningen on the topic of artificial intelligence and hyperspectral imaging.
\end{IEEEbiography}

\begin{IEEEbiography}[{\includegraphics[width=1in,height=1in,clip,keepaspectratio]{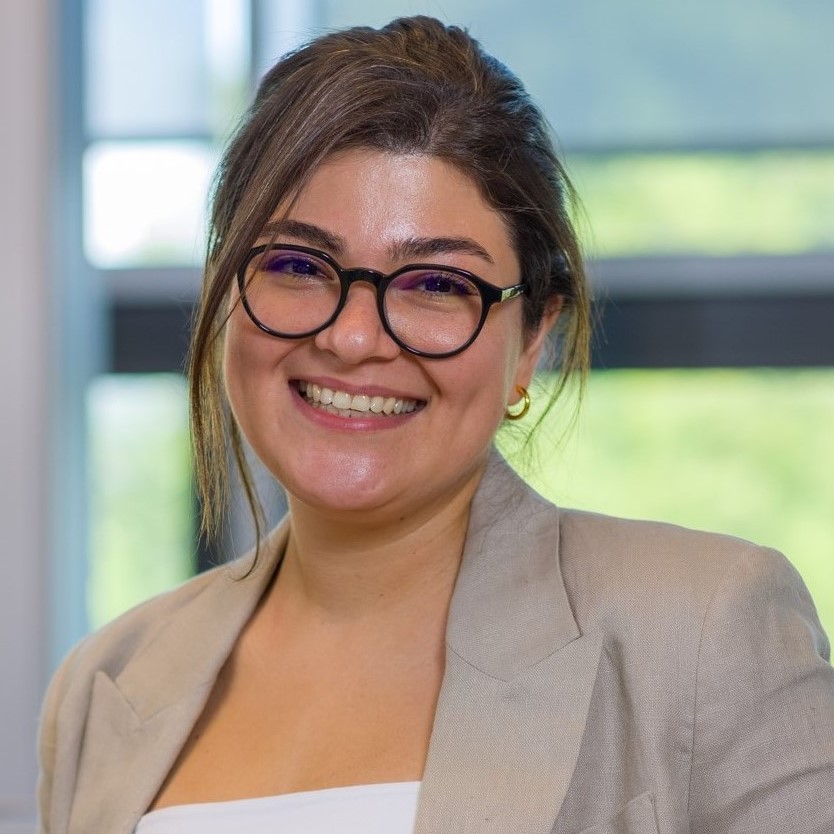}}]{Maya Aghaei} is a researcher in Computer Vision and
Data Science research group at the NHL Stenden University
of Applied Sciences in Leeuwarden, The Netherlands. Supported by the history of working in industrial research, her main research interest is in the field of applied computer vision and machine learning. With a background in computer software engineering, she obtained her PhD focused on computer vision applied to social signal processing from University of Barcelona, Spain, in 2018, and consequently spent over two years of as a postdoctoral researcher at Italian Institute of Technology, Genova, Italy.
\end{IEEEbiography}

\begin{sloppypar}
\begin{IEEEbiography}[{\includegraphics[width=1in,height=1in,clip,keepaspectratio]{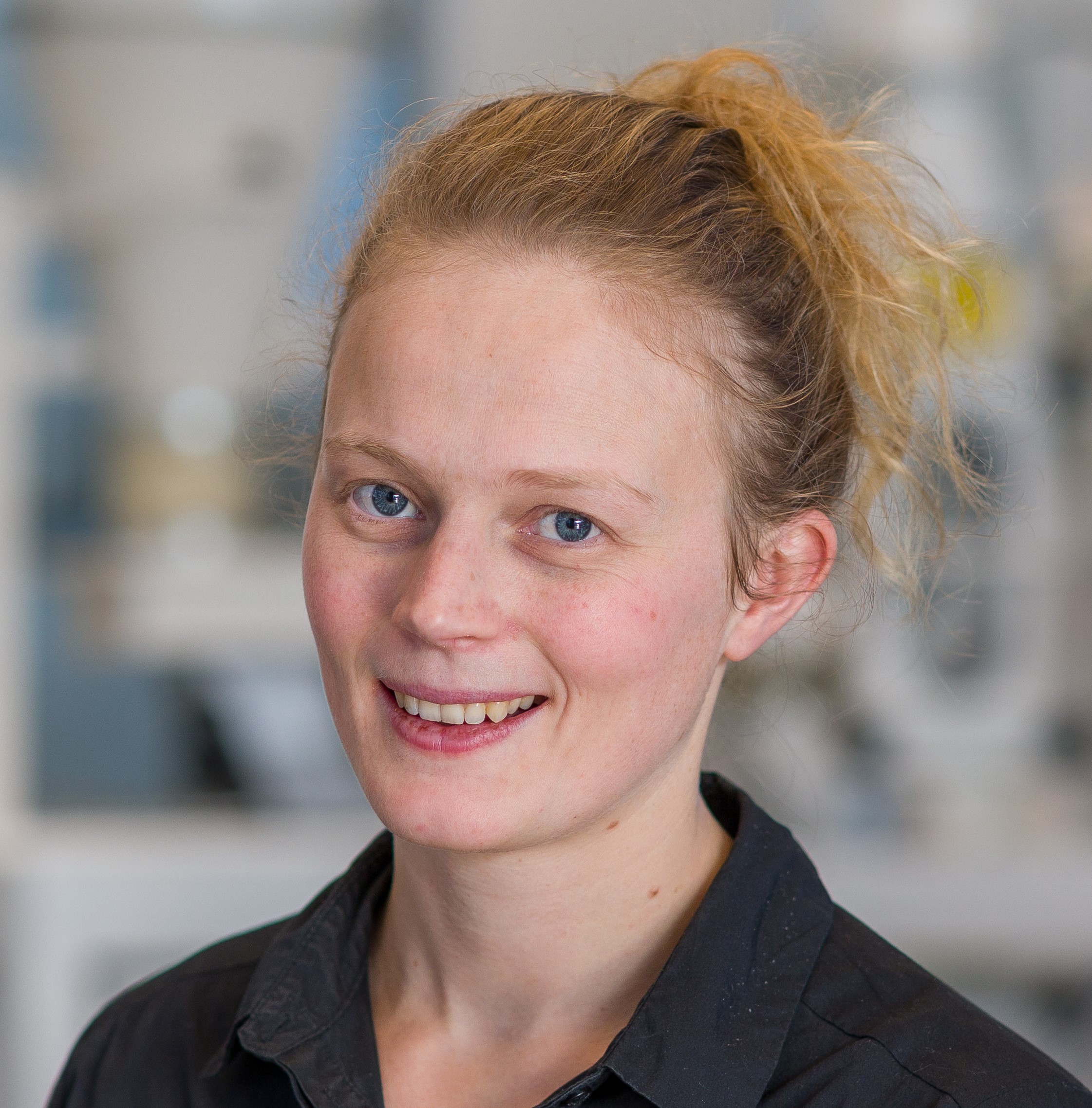}}]{Femke Jaarsma} is researcher at the professorship Circular Plastics of NHL Stenden University of Applied Sciences in Leeuwarden. In her research, she aims to improve the process of mechanical recycling of plastic waste streams by optimised sorting and washing. Before her work at the professorship Circular Plastics she obtained her B.Sc. and M.Sc. degree at Wageningen University and Research, where she studied Biotechnology with specialisations Environmental and Biobased Technology and Process Technology.
\end{IEEEbiography}
\end{sloppypar}

\begin{sloppypar}
\begin{IEEEbiography}[{\includegraphics[width=1in,height=1in,clip,keepaspectratio]{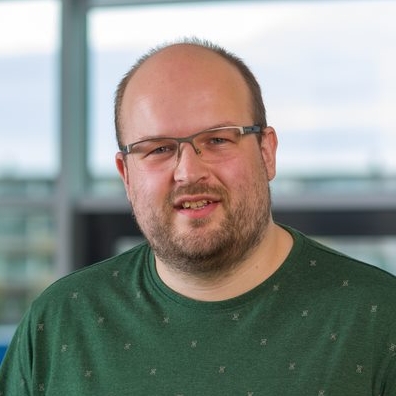}}]{Martin Dijkstra} is an engineer at the Professorship Computer Vision \& Data Science of the NHL Stenden University of Applied Sciences in Leeuwarden. His interests are in computer vision, deep learning, (hyperspectral) camera technology, and the software development involved therewith. He became involved with the professorship during his Software Engineering study and has continued his work there after obtaining his B.Eng. degree in that field in 2014.
\end{IEEEbiography}
\end{sloppypar}

\begin{IEEEbiography}[{\includegraphics[width=1in,height=1in,clip,keepaspectratio]{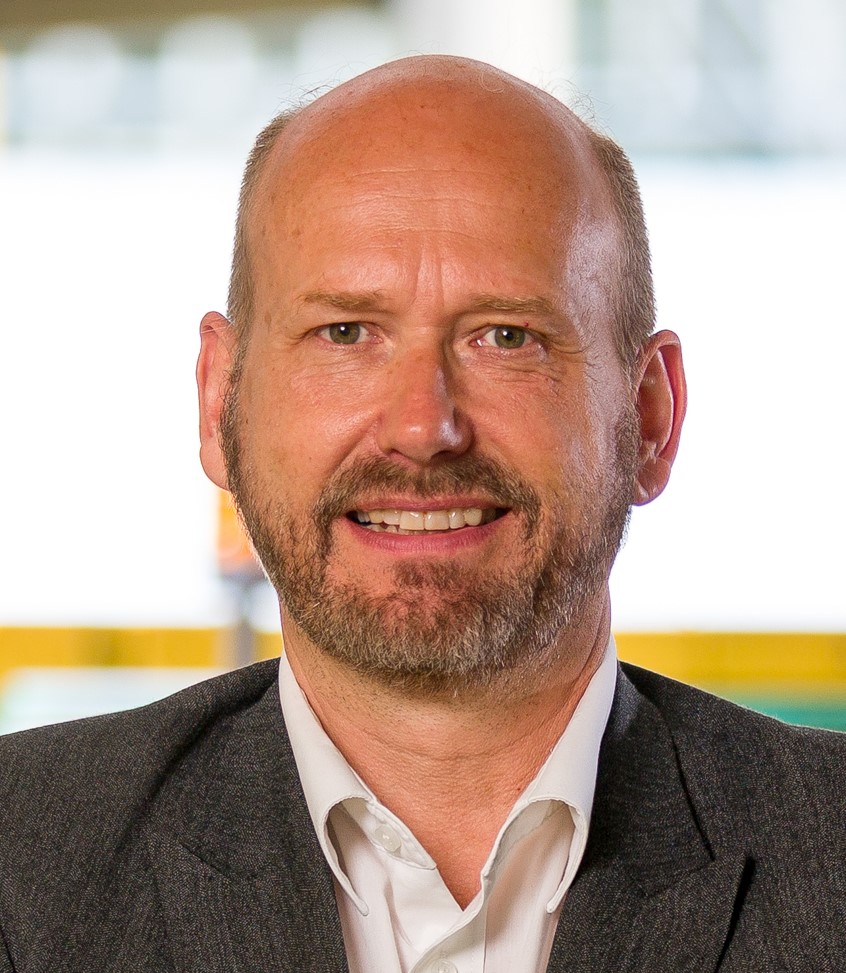}}]{Rudy Folkersma} obtained his MSc in Polymer Chemistry from the University of Groningen in 1990, and went on to obtain his PhD with his thesis “Microgravity Coagulation and Particle Gel Formation” at Eindhoven University of Technology.
From 1997 to 2000 Rudy worked at the Dutch Polymer Institute (DPI) in Eindhoven as a research associate. As a Professor of Applied Sciences, Rudy started at NHL Stenden in 2011 focusing on innovations in the field of sustainable plastics technology in cooperation with the business community. Apart from being a professor at the professorship Sustainable Plastics he also joined in 2016 the professorship Circular Plastics. In addition, he developed two minors and a Master program in the field of biobased and circular plastics.
\end{IEEEbiography}

\begin{IEEEbiography}[{\includegraphics[width=1in,height=1in,clip,keepaspectratio]{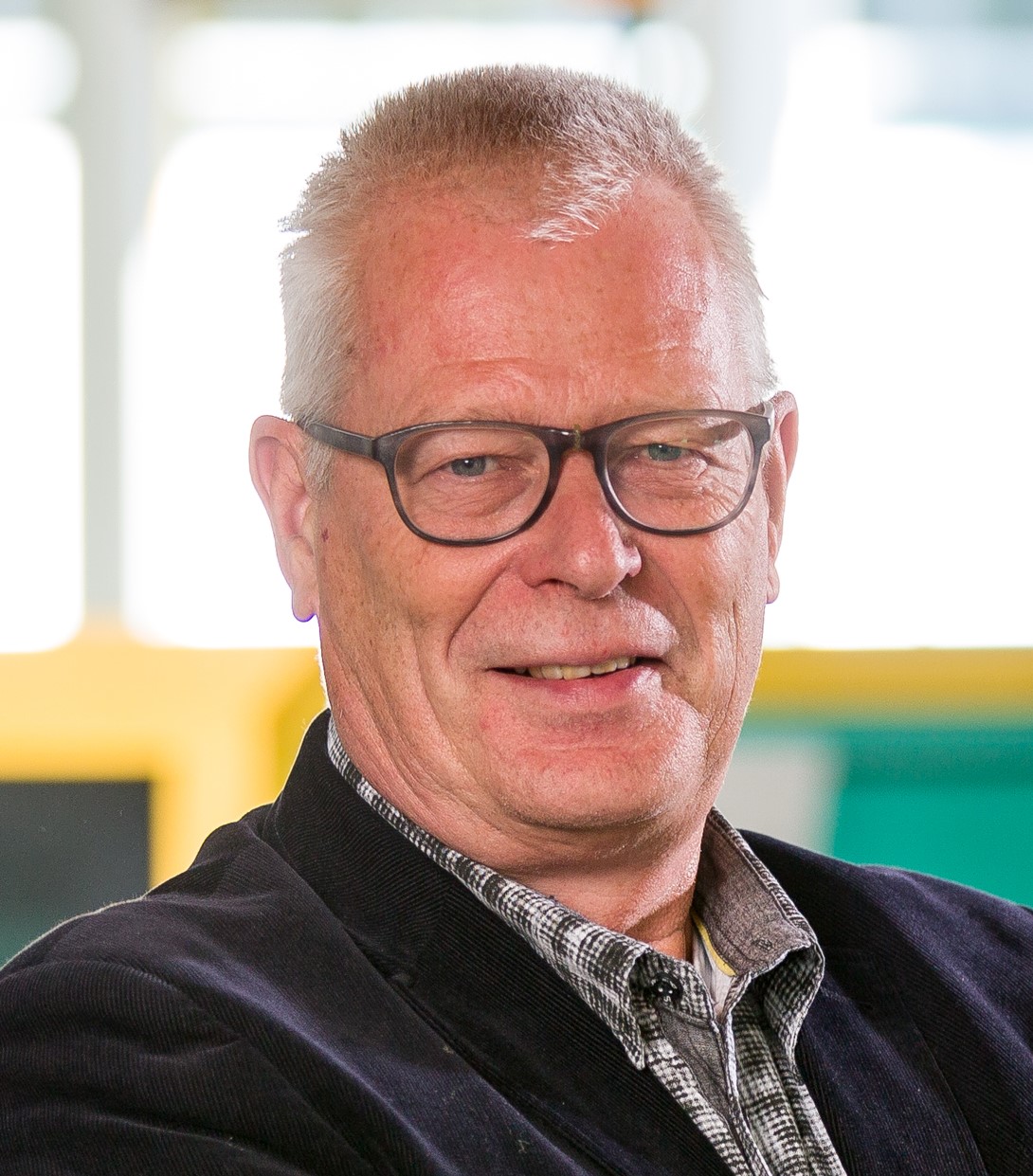}}]{Jan Jager} studied and did his PhD in physical organic research at the University of Groningen before joining Akzo Nobel in Arnhem where he spent several years conducting research into various plastics for new textile developments. After the winding-up of Diolen Industrial Fibers in Emmen he became co-founder of Applied Polymer Innovations (API, now Senbis Polymer Innovations), which emerged from Diolen’s former R\&D department. In April 2011 Jan gained an appointment as part-time Professor at NHL Stenden University of Applied Sciences. Within NHL Stenden, Jan is now full-time involved in research and education on upcycling and recycling of plastics (Professorship Circular Plastics) and the development of new products based on biopolymers, biobased and biodegradables (Professorship Sustainable Polymers).
\end{IEEEbiography}

\begin{sloppypar}
\begin{IEEEbiography}[{\includegraphics[width=1in,height=1in,clip,keepaspectratio]{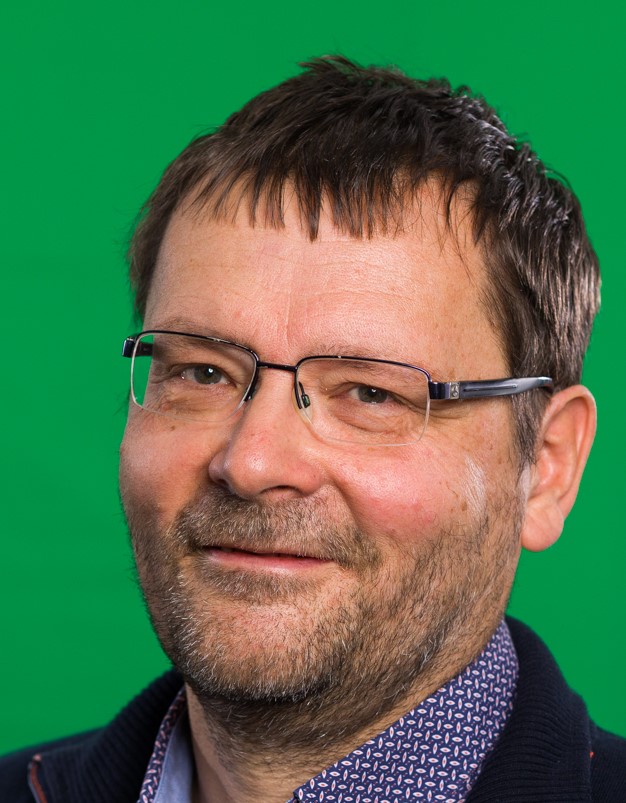}}]{Jaap van de Loosdrecht} is a professor of applied sciences in Computer Vision Data Science at the NHL Stenden University of Applied Sciences in Leeuwarden. His main research interests are in computer vision, deep learning and hyperspectral imaging. He has been active in the field of computer vision since 1996.
\end{IEEEbiography}
\end{sloppypar}







\end{document}